\documentclass[10pt,journal,compsoc]{IEEEtran}
\usepackage{graphicx}
\usepackage{amsfonts}

\usepackage[english]{babel}
\usepackage{amsmath, amssymb}
\usepackage{float}
\usepackage{makecell}
\usepackage{multirow}
\usepackage{amsmath,amssymb,amsfonts}
\usepackage{algorithmic}
\usepackage{graphicx}
\usepackage{textcomp}
\usepackage{xcolor}

\usepackage{subfigure}
\usepackage{epstopdf}

\usepackage[pagebackref=false,breaklinks=true,citecolor=black,linkcolor=black,letterpaper=true,colorlinks,bookmarks=false]{hyperref}

\usepackage{graphicx}
\usepackage{comment}
\usepackage{amsmath,amssymb} 
\usepackage{color}
\usepackage{enumerate}

\usepackage{booktabs}

\usepackage{soul}
\soulregister\cite7 
\soulregister\citep7
\soulregister\citet7
\soulregister\ref7 
\soulregister\pageref7

\ifCLASSOPTIONcompsoc
  \usepackage[nocompress]{cite}
\else
  \usepackage{cite}
\fi
\ifCLASSINFOpdf
\else
\fi
\begin{document}
%
\title{Efficient 3D Deep LiDAR Odometry}
%
%
%
%

\author{Guangming Wang, Xinrui Wu, Shuyang Jiang, Zhe Liu,
        and Hesheng Wang*
\IEEEcompsocitemizethanks{\IEEEcompsocthanksitem G. Wang, X. Wu, and H. Wang are with Department of Automation, Key Laboratory of System Control and Information Processing of Ministry of Education, Key Laboratory of Marine Intelligent Equipment and System of Ministry of Education, Shanghai Engineering Research Center of Intelligent Control and Management, Shanghai Jiao Tong University, Shanghai 200240, China. S. Jiang is with Department of Computer Science and Engineering, Shanghai Jiao Tong University, Shanghai 200240, China. Z. Liu is with the Key Laboratory of Artificial Intelligence of Ministry of Education, Shanghai Jiao Tong University, Shanghai 200240, China.\protect
\IEEEcompsocthanksitem  The first two authors contributed equally to this work. Corresponding Author: Hesheng Wang (wanghesheng@sjtu.edu.cn).}
}

%
%

\markboth{Journal of \LaTeX\ Class Files,~Vol.~14, No.~8, August~2015}%
{Shell \MakeLowercase{\textit{et al.}}: Bare Demo of IEEEtran.cls for Computer Society Journals}
%



\IEEEtitleabstractindextext{%
\begin{abstract}
 An efficient 3D point cloud learning architecture, named EfficientLO-Net, for LiDAR odometry is first proposed in this paper. In this architecture, the projection-aware representation of the 3D point cloud is proposed to organize the raw 3D point cloud into an ordered data form to achieve efficiency. The Pyramid, Warping, and Cost volume (PWC) structure for the LiDAR odometry task is built to estimate and refine the pose in a coarse-to-fine approach. A projection-aware attentive cost volume is built to directly associate two discrete point clouds and obtain embedding motion patterns. Then, a trainable embedding mask is proposed to weigh the local motion patterns to regress the overall pose and filter outlier points. The trainable pose warp-refinement module is iteratively used with embedding mask optimized hierarchically to make the pose estimation more robust for outliers. The entire architecture is holistically optimized end-to-end to achieve adaptive learning of cost volume and mask, and all operations involving point cloud sampling and grouping are accelerated by projection-aware 3D feature learning methods. The superior performance and effectiveness of our LiDAR odometry architecture are demonstrated on KITTI, M2DGR, and Argoverse datasets. Our method outperforms all recent learning-based methods and even the geometry-based approach, LOAM with mapping optimization, on most sequences of KITTI odometry dataset. We open sourced our codes at: https://github.com/IRMVLab/EfficientLO-Net.
\end{abstract}

}

\maketitle

\IEEEdisplaynontitleabstractindextext

%
\IEEEpeerreviewmaketitle

\IEEEraisesectionheading{\section{Introduction}\label{sec:introduction}}

\IEEEPARstart{V}{isual}/LiDAR odometry is one of the key technologies in autonomous driving, and acts as the base of the subsequential planning and decision making of mobile robots. This task uses two consecutive images or point clouds to obtain the relative pose transformation between two frames. Recently, learning-based odometry methods have shown impressive accuracy on datasets compared with conventional methods based on hand-crafted features. It is found that learning-based methods can deal with sparse features and dynamic environments \cite{dasgil,lpdnet}, which are usually difficult for conventional methods. To our knowledge, most learning-based methods are on the 2D visual odometry \cite{wang2017deepvo,zhou2017unsupervised,Unsupervised_depth,chizhang,shamwell2019unsupervised,min2020voldor,yang2020d3vo} or utilize 2D convolution on the projected information of LiDAR \cite{nicolai2016deep,velas2018cnn,wang2019deeppco,li2019net,li2020dmlo}. 3D learning methods from the raw point cloud have been developed rapidly and have recently made remarkable progress on many problem \cite{qi2017pointnet,qi2017pointnet++,qi2020p2b,liu2019flownet3d,liu2019meteornet}, while the deep LiDAR odometry in 3D point clouds is underexplored. This paper aims to learning the LiDAR odometry directly through 3D point clouds.

\begin{figure}[t]
	\centering
	\resizebox{0.90\columnwidth}{!}
	{
		\includegraphics[scale=1.0]{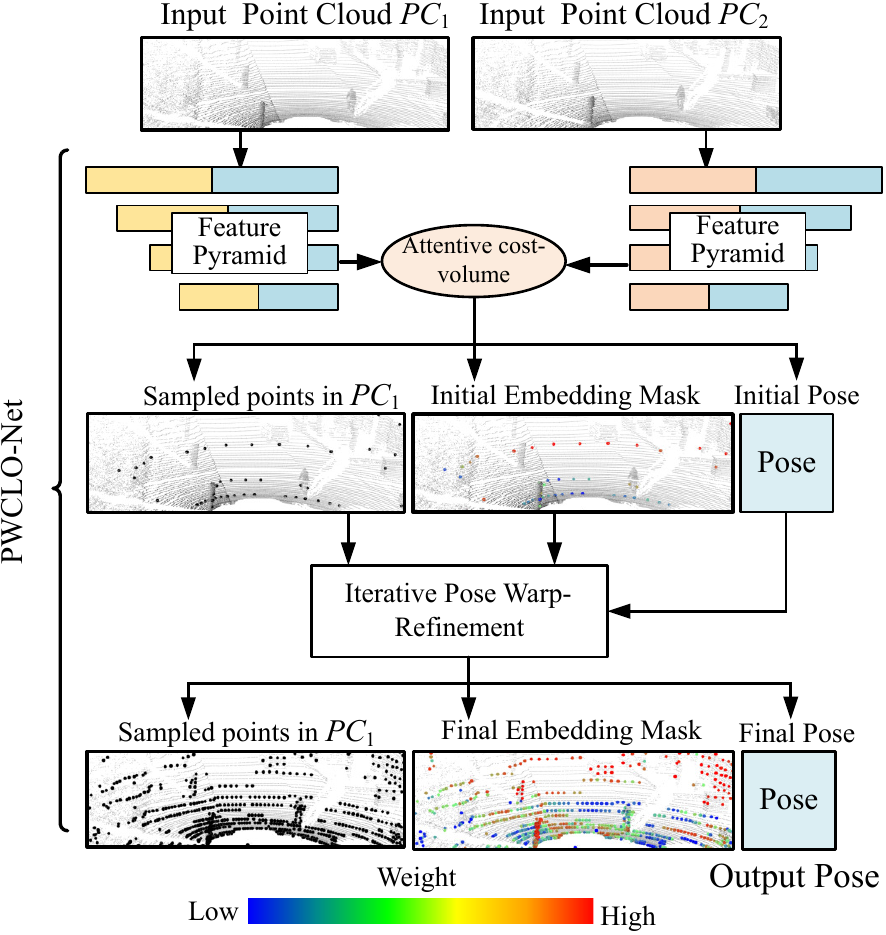}}
	\vspace{-2mm}
	\caption{The Point Feature Pyramid, Pose Warping, and Attentive Cost Volume (PWC) structure in our proposed EfficientLO-Net. The pose is refined layer by layer through iterative pose warp-refinement. The whole process is realized end-to-end by making all modules differentiable. In the LiDAR point clouds, the small black points are the whole point cloud. The big black points are the sampled points in $PC_1$. Distinctive colors of big points in embedding mask measure the contribution of sampled points to the pose estimation.}
	\label{fig:network_little}
	\vspace{-0pt}
\end{figure}

For the learning of LiDAR odometry from 3D point clouds, there are four challenges: 1) As the discrete LiDAR point data are obtained in two consecutive frames separately, it is hard to find a precise corresponding point pair between two frames; 2) Some points belonging to an object in a frame may not be seen in the other view if they are occluded by other objects or are not captured because of the limitation of LiDAR resolution; 3) Points belonging to dynamic objects are not suitable to be used for the pose estimation because these points have uncertainty motions of the dynamic objects; 4) Learning from the raw 3D point cloud is often troubled by lower efficiency because of the irregularity and disorder of the raw point cloud.

For the first challenge, Zheng et al. \cite{zheng2020lodonet} use matched keypoint pairs judged in 2D depth images. However, the correspondence is rough because of the discrete perception of LiDAR. The cost volume for 3D point cloud \cite{wu2019pointpwc,wang2021hierarchical} is adopted in this paper to obtain weighted soft correspondence between two consecutive frames.
We directly regress overall pose from the cost volume containing soft correspondence information, instead of calculating overall pose from estimated explicit corresponding points \cite{wang2021hierarchical}. We argue that omitting the unnecessary step of corresponding point estimation can reduce the difficulty of learning and bring better learning results. This is demonstrated in this paper through experiments. 

 For the second and third challenges, the mismatched points or dynamic objects, which do not conform to the overall pose, need to be filtered. LO-Net~\cite{li2019net} trains an extra mask estimation network~\cite{zhou2017unsupervised, yang2018unsupervised} by minimizing the consistency error of the normal of 3D points. In our network, an internal trainable embedding mask is proposed to weigh local motion patterns from the cost volume to regress the overall pose. In this way, the mask can be optimized for more accurate pose estimation rather than depending on geometry correspondence. In addition, the PWC structure is built to capture the large motion in the layer of sparse points and refine the estimated pose in dense layers. As shown in Fig.~\ref{fig:network_little}, the embedding mask is also optimized hierarchically to obtain more accurate filtering information to refine pose estimation.

For the fourth challenge, efficiently organizing the 3D point cloud for feature learning is the key to solving the problem. As mentioned in recent studies \cite{rethage2018fully,liu2019point,hu2020randla,xu2020grid,hu2021learning}, the commonly used Farthest Point Sampling (FPS)~\cite{qi2017pointnet++} and $K$ Nearest Neighbors (KNN)~\cite{qi2017pointnet++} are inefficient and even unreasonable \cite{hermosilla2018monte} because of the uneven density of 3D point cloud. Specifically, FPS defines the center position for grouping, and KNN groups raw disordered point cloud for feature aggregation. They all need to access the entire point cloud multiple times, which requires a huge amount of calculation. Random sampling \cite{hu2020randla,hu2021learning} and voxelization-based method \cite{riegler2017octnet,rethage2018fully} are criticized because of the uneven density of the point cloud \cite{xu2020grid} and the information loss from quantization error \cite{liu2019point}, respectively. {According to the line scanning characteristic of mechanical LiDAR, points are usually closer in 3D space when they are closer in cylindrical projection space, except for the edges of the foreground and background. Based on this characteristic, we propose to organize the 3D point cloud in the cylindrical projection space for efficient feature learning. Specifically, we propose a projection-aware representation of 3D point cloud, and the corresponding feature learning methods, including stride-based sampling strategy, the efficient projection-aware grouping strategy, and the 3D distance-based filtering strategy. These form the framework of the projection-aware 3D feature learning method and are applied to the efficient 3D deep LiDAR odometry task.}

Overall, our contributions are as follows:
\begin{itemize}
	\item We build a totally end-to-end  efficient framework, {named EfficientLO-Net,} for the 3D LiDAR odometry task, in which all modules are fully differentiable so that each process is no longer independent but holistically optimized. {Combined with the characteristics of mechanical LiDAR scanning, a projection-aware representation of 3D point cloud and the corresponding feature learning methods, including stride-based sampling strategy, the efficient projection-aware grouping strategy, and the 3D distance-based filtering strategy are proposed to organize the unstructured 3D point cloud data into an ordered form and achieve efficient learning of large-scale point clouds.}
	\item In the proposed framework, the hierarchical embedding mask is proposed to filter mismatched points and convert the cost volume embedded in points to the overall pose in each refinement level. Meanwhile, the embedding mask is optimized and refined hierarchically to obtain more accurate filtering information for pose refinement following the density of points.
	\item Based on the characteristic of the pose transformation, the pose warping and pose refinement are proposed to refine the estimated pose layer by layer iteratively. {The projection-aware 3D feature learning methods are integrated into all operations involving point cloud sampling and grouping to speed up the 3D LiDAR odometry.}
	\item   The evaluation experiments and ablation studies on KITTI odometry dataset \cite{geiger2012we,geiger2013vision} show the superiority of the proposed method and the effectiveness of each design. To the best of our knowledge, our method outperforms all recent learning-based LiDAR odometry and even outperforms the geometry-based LOAM with mapping optimization \cite{zhang2017low} on most sequences. 
\end{itemize}

\section{Related Work}
\subsection{Efficient 3D point cloud Learning}

PointNet \cite{qi2017pointnet} proposes the feature learning method for the raw unstructured 3D point cloud. Later, the hierarchical feature extraction method for raw 3D point cloud is proposed by PointNet++ \cite{qi2017pointnet++}, which groups points and extracts 3D local features hierarchically from 3D point cloud through the FPS and KNN or ball query operator, and improves the performance of point cloud learning. However, the grouping process of the disordered, unstructured, and uneven 3D point cloud also brings more calculations. This makes many studies only consume 8192 or fewer points in various 3D point cloud based tasks, including 3D object tracking \cite{qi2020p2b}, 3D classification \cite{mao2019interpolated,liu2019meteornet,9522122,xu2020grid}, 3D semantic segmentation \cite{wu2019pointconv,thomas2019kpconv,lin2020fpconv}, 3D scene flow \cite{liu2019flownet3d,wu2019pointpwc,wang2021hierarchical}, and 3D LiDAR odometry \cite{wang2021pwclo}. There are a series of works trying to solve this problem. HPLFlowNet \cite{gu2019hplflownet} uses lattice-based interpolation to perform the inference on large-scale point clouds. However, the interpolation brings errors. Voxel-based methods have been studied a lot, but the amount of memory and calculation increases dramatically with the growth of voxel resolution. OctNet \cite{riegler2017octnet} is proposed to solve the problem of memory, but the quantization error caused by voxelization is inevitable. PVCNN \cite{liu2019point} does not sample the point cloud, but instead combines the voxel with individual points to accelerate the learning process for 3D point cloud. However, the points in a voxel grid are treated indiscriminately, which makes the feature learning in the local point cloud block  still have quantization errors. RandLA-Net \cite{hu2020randla,hu2021learning} uses random sampling to improve the efficiency of selecting center points. However, random sampling will cause insufficient coverage of the 3D space due to the uneven density of the point cloud \cite{xu2020grid}. A new local feature aggregation method based on local spatial encoding and attentive pooling is proposed to alleviate this shortcoming, so that large-scale 3D point clouds can be processed and the method gets a promising result. Grid-GCN \cite{xu2020grid} stores multiple raw point coordinates in each voxel, and proposes corresponding sampling strategies and feature extraction methods for this data structure. Although storing multiple points in each voxel reduces the quantization error of one voxel taking one point in voxelization, each voxel grid needs to retain many points to retain all raw points, which leads to a linear increase of the memory and the amount of calculation. With the expansion of the scene and the aggravation of uneven point cloud density, like the large-scale LiDAR point cloud, the number of points located in voxels will become more uneven. It becomes more difficult to balance the voxel resolution and the number of points saved in each voxel, which makes it difficult for Grid-GCN to process large-scale LiDAR datasets.

This paper proposes a novel projection-aware 3D feature learning method inspired from the line scan characteristics of LiDAR. In this method, by taking advantage of the characteristics of LiDAR data itself, classic point cloud feature learning and inter-frame feature correlation can be accelerated, avoiding the use of 3D voxels and complex data structure design like Grid-GCN \cite{xu2020grid}. Meanwhile, raw 3D point clouds are retained to avoid loss of information. Based on these, we realize a {20 Hz} real-time 3D deep LiDAR odometry, and achieve higher accuracy due to the use of large-scale LiDAR point clouds compared to our previous work \cite{wang2021pwclo}, which uses 8192 points for each frame.

\subsection{Deep LiDAR Odometry}

Deep learning has gained impressive progress in visual odometry \cite{min2020voldor,yang2020d3vo}. However, the 3D LiDAR odometry with deep learning is still a challenging problem. In the beginning, Nicolai et al. \cite{nicolai2016deep} project two consecutive LiDAR point clouds to the 2D plane to obtain two 2D depth images, and then use the 2D convolution and fully connected (FC) layers to realize learning-based LiDAR odometry. Their work verifies that the learning-based method is serviceable for the LiDAR odometry although their experiment results are not superior. Velas et al. \cite{velas2018cnn} also project LiDAR points to the 2D plane but use three channels to encode the information, including height, range, and intensity. Then the convolution and FC layers are applied for the pose regression. The performance is superb when only estimating the translation but is poor when estimating the 6-DOF pose. Wang et al. \cite{wang2019deeppco} project point clouds to panoramic depth images and stack two frames together as input. Then the translation sub-network and FlowNet \cite{dosovitskiy2015flownet} orientation sub-network are employed to estimate the translation and orientation respectively. Li et al. \cite{li2019net} also preprocess 3D LiDAR points to 2D information but use the cylindrical projection \cite{chen2017multi}. Then, the normal of each 3D point is estimated to build consistency constraint between frames, and an uncertainty mask is estimated to mask dynamic regions. Zheng et al. \cite{zheng2020lodonet} extract matched keypoint pairs by classic detecting and matching algorithms in 2D spherical depth images projected from 3D LiDAR point clouds. Then the PointNet-based structure is used to regress the pose from the matched keypoint pairs. \cite{li2020dmlo} proposes a learning-based network to generate matching pairs with high confidence, then applies Singular Value Decomposition (SVD) to get the 6-DoF pose. \cite{cho2020} introduces an unsupervised learning method on LiDAR odometry.

\begin{figure*}[t]
	\centering
	\resizebox{1.00\textwidth}{!}
	{
		\includegraphics[scale=1.00]{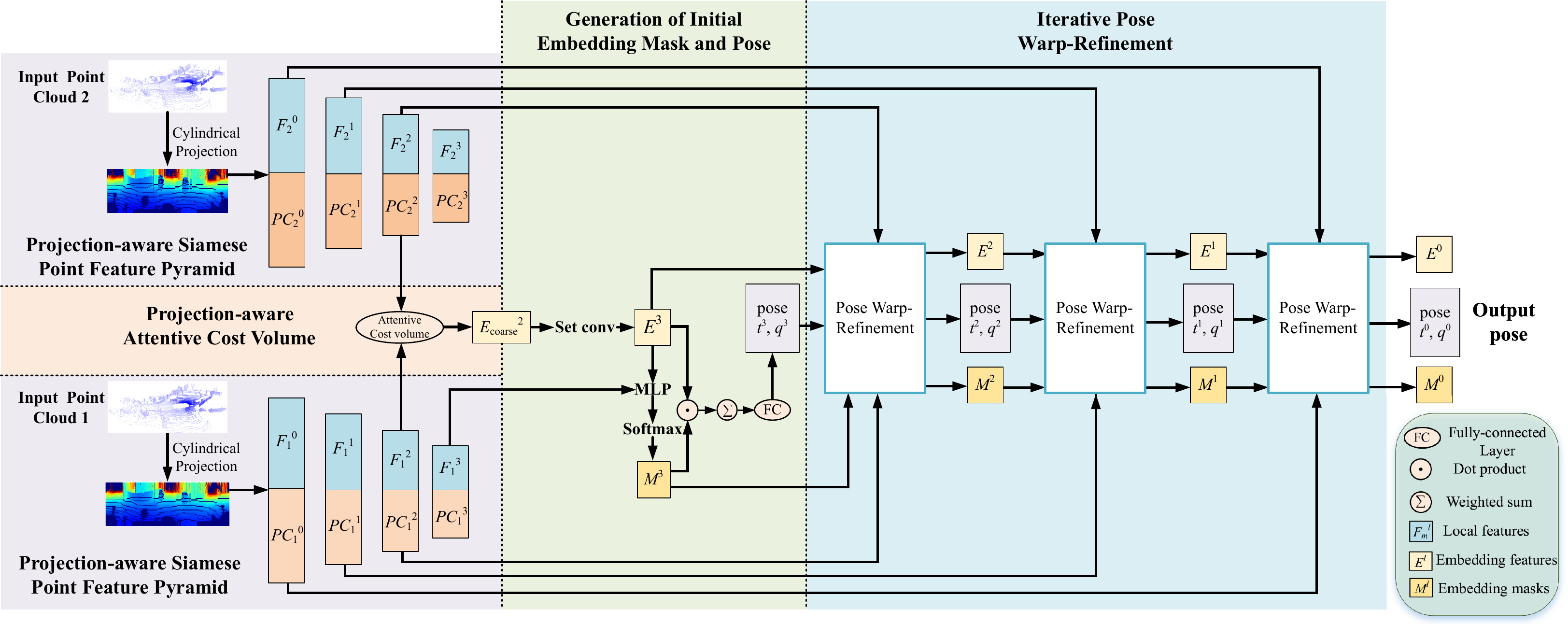}}
	\vspace{-7mm}
	\caption{The details of the proposed EfficientLO-Net architecture. The network is composed of four set conv layers in point feature pyramid, one attentive cost volume, one initial embedding mask and pose generation module, and three pose warp-refinement modules. The network outputs the predicted poses from four levels for supervised training. }
	\label{fig:network}
	\vspace{-2pt}
\end{figure*}

\begin{figure}[t]
	\centering
	\resizebox{1.0\columnwidth}{!}
	{
		\includegraphics[scale=1.00]{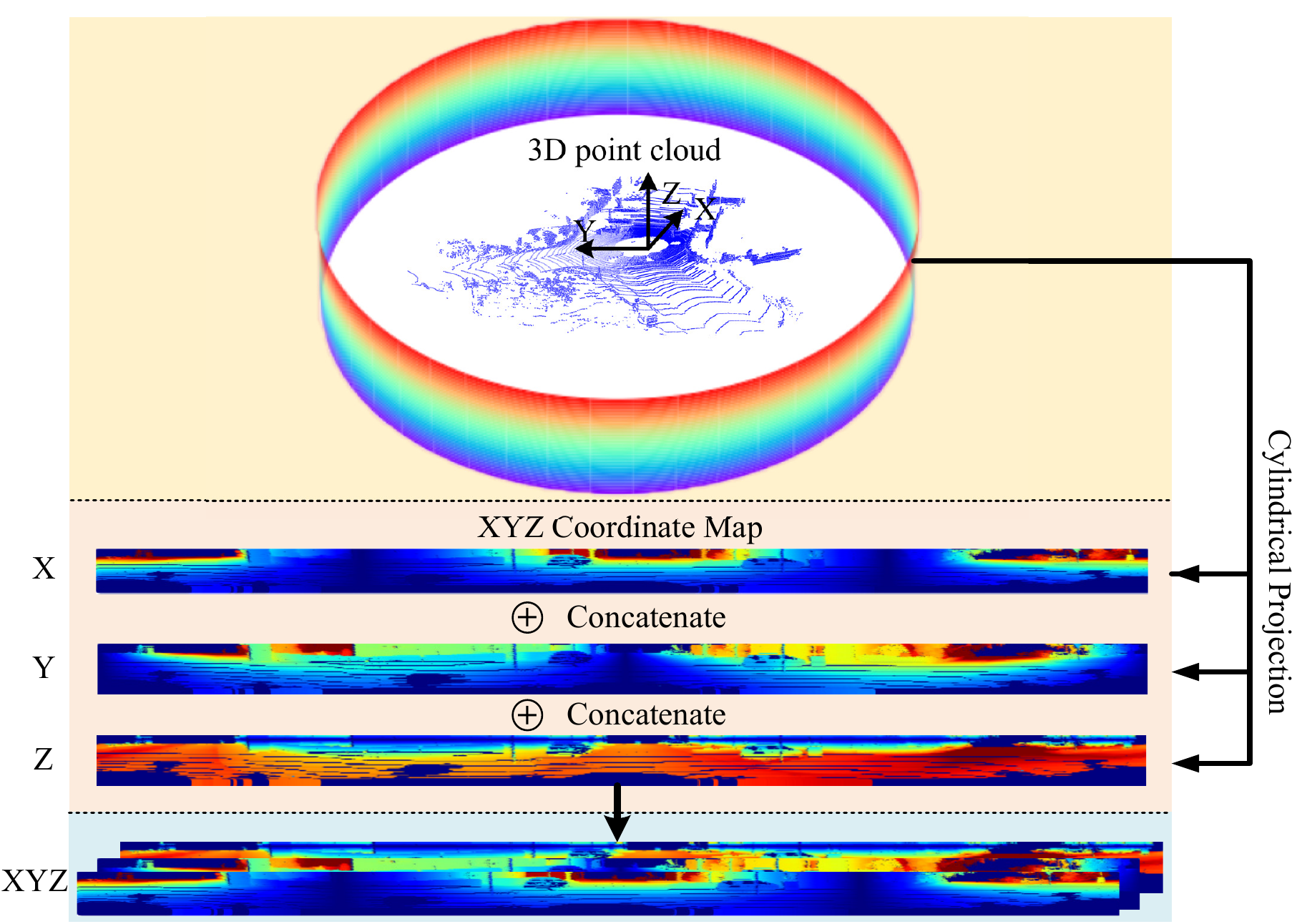}}
	\vspace{-4mm}
	\caption{Visualization of the projection-aware representation of point clouds. During the projection process, the 3D point cloud will be projected onto a cylindrical plane according to the parameters of the LiDAR. The raw XYZ coordinates will be recorded in the cylindrical plane. For subsequent operations, the left and right sides of the XYZ coordinate map obtained from cylindrical projection are connected.
}
	\label{fig:projection}
\end{figure}

\subsection{Deep Point Correlation}

The above studies all use the 2D projection information for the LiDAR odometry learning, which converts the LiDAR odometry to the 2D learning problem. Wang et al. \cite{wang2019deeppco} compare the 3D point input and 2D projection information input based on the same 2D convolution model. It is found that the 3D input based method has a poor performance. As the development of 3D deep learning \cite{qi2017pointnet,qi2017pointnet++}, FlowNet3D \cite{liu2019flownet3d} proposes an embedding layer to learn the correlation between the points in two consecutive frames. After that, Wu et al. \cite{wu2019pointpwc} propose the cost volume method on point clouds, and HALFlow \cite{wang2021hierarchical} develop the attentive cost volume method. The point cost volume involves the motion patterns of each point. It becomes a new direction and challenge to regress pose from the cost volume, and meanwhile, not all point motions are for the overall pose motion. We explore to estimate the pose directly from raw 3D point cloud data and deal with the new challenges encountered.

Moreover, we are inspired by the Pyramid, Warping, and Cost volume (PWC) structure in the flow network proposed by Su et al. \cite{sun2018pwc,sun2019models}. The work uses three modules (Pyramid, Warping, and Cost volume) to refine the optical flow through a coarse-to-fine approach. The works \cite{wu2019pointpwc, wang2021hierarchical} on 3D scene flow also use the PWC structure to refine the estimated 3D scene flow in point clouds. The trainable pose refinement in this paper is inspired from this, and a PWC structure for LiDAR odometry is built for the first time.

\section{Projection-aware EfficientLO-Net}

\subsection{Overview and Projection-aware Representation of 3D Point Clouds}\label{sec:Representation}
{The mechanical LiDAR sensors, such as Velodyne and Ouster, generate a view of the complete environment by rotating multiple LiDAR emitters and photon detectors at high frequency to generate 3D point measures. Therefore, the mechanical LiDAR sensors generate range images which is usually then converted into 3D point clouds. As LiDAR datasets, like KITTI, provide only 3D point clouds, where the motion while scanning is already accounted for, previous methods \cite{nicolai2016deep,velas2018cnn,wang2019deeppco,li2019net,li2020dmlo} project the LiDAR point cloud onto a cylindrical surface to obtain a 2D range picture.} Then, the classic 2D convolution can be used with high efficiency. However, this ignores the 3D characteristic of LiDAR point clouds.
{Many recent studies \cite{qi2017pointnet, qi2017pointnet++, hu2020randla, liu2019flownet3d, lpdnet, qi2020p2b} have shown the advantages of 3D point cloud learning method for 3D tasks.} The 6-DOF pose transformation estimation for the odometry is also a 3D task. For this task, the registration and pose regression in 3D space are more direct and explicit. Therefore, in our architecture, it is expected to utilize the advantages of 3D point cloud learning method, but also accelerate the learning efficiency of the large-scale 3D point cloud. Due to the occlusion, a single-frame LiDAR point cloud is only distributed on the surface of objects. Using this prior knowledge, the raw disordered point cloud can be organized orderly according to the proximity relationship of the point cloud to facilitate the commonly used point cloud grouping operations. Therefore, we propose the projection-aware representation of 3D point clouds to {organize} unstructured 3D data to orderly form by the cylindrical projection prior. In the process of cylindrical projection, the 3D point cloud will be kept on the 2D cylindrical surface orderly, and then the corresponding projection-aware 3D learning operations are proposed to efficiently process the raw 3D coordinates.

\begin{figure*}[t]
	\centering
	\resizebox{1.00\textwidth}{!}
	{
		\includegraphics[scale=1.00]{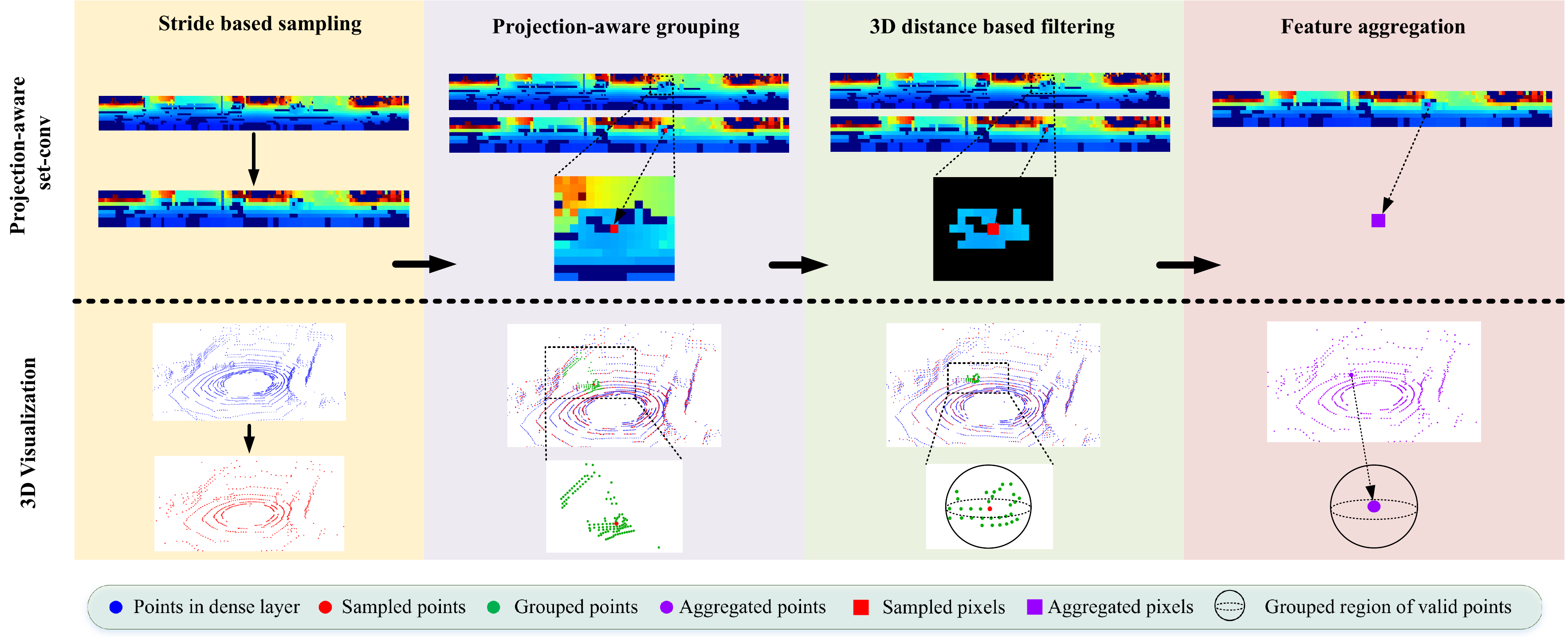}}
	\vspace{-6mm}
	\caption{The details of projection-aware set-conv modules. The module is composed of four key parts: stride-based sampling, projection-aware grouping, 3D distance-based filtering, and feature aggregation. The output is the sampled points with aggregated feature on the corresponding projected 2D plane.}
	\label{fig:new_ops1}
\end{figure*}

Fig.~\ref{fig:network} illustrates the overall structure of our EfficientLO-Net. The inputs to the network are two point clouds $PC_1 = \{ {x_i}|{x_i} \in {\mathbb{R}^3}\} _{i = 1}^{N_1}$ and $PC_2 = \{ {y_j}|{y_j} \in {\mathbb{R}^3}\} _{j = 1}^{N_2}$ 
from two adjacent frames. 
First, the raw 3D points are projected onto a cylindrical plane to obtain the 2D coordinates corresponding to each 3D point. The formula is as follows:
\begin{equation}
	u = \left\lfloor {\arctan 2(y/x)/\Delta \theta } \right\rfloor,  
\end{equation}
\begin{equation}
	v = \left\lfloor {\arcsin (z/\sqrt {{x^2} + {y^2} + {z^2}} )/\Delta \phi } \right\rfloor.  
\end{equation} 
Previous projection-based methods fill in the projected 2D coordinate position with the depth values \cite{nicolai2016deep,wang2019deeppco,zheng2020lodonet}, the range and intensity values\cite{velas2018cnn}, the normal values\cite{li2019net}, the range and reflectivity values\cite{li2020dmlo}, and use 2D convolution network to learning LiDAR odometry from these.
{Different from these learning-based LiDAR odometry methods, we fill in the projected 2D coordinate position with the raw 3D point cloud coordinates as shown in Fig. \ref{fig:projection}, like the data preprocessing methods of the geometry-based SLAM method, SuMa \cite{behley2018efficient}, and some 3D semantic segmentation methods \cite{milioto2019rangenet++,xu2020squeezesegv3,kochanov2020kprnet,cortinhal2020salsanext}.} The input to our network is a tensor of $H \times W \times 3$ with XYZ 3D coordinates. Because the XYZ coordinate map is only a new orderly data organization form of 3D point clouds, we refer it as projection-aware representation of point clouds. Our network still consumes raw 3D point cloud coordinates and {performs} 3D feature learning in Euclidean space, while other methods \cite{nicolai2016deep,wang2019deeppco,zheng2020lodonet,velas2018cnn,li2019net,li2020dmlo} use 2D convolution for the feature learning of projected data. 

As shown in Fig.~\ref{fig:network}, the new projection-aware representations of 3D point clouds are firstly encoded by the projection-aware siamese feature pyramid consisting of several projection-aware set conv layers as introduced in Sec.~\ref{sec:pyramid}. Then the projection-aware attentive cost volume is proposed to generate embedding features, which will be described in Sec.~\ref{sec:cost}. To regress pose transformation from the embedding features, hierarchical embedding mask optimization is proposed in Sec.~\ref{sec:mask}. Next, the pose warp-refinement method is proposed in Sec.~\ref{sec:refine} to refine the pose estimation in a coarse-to-fine approach. Finally, the network outputs the quaternion $q \in {\mathbb{R}^4}$ and translation vector $t \in {\mathbb{R}^3}$.

\subsection{Projection-aware Siamese Point Feature Pyramid}\label{sec:pyramid}
The input point clouds are usually disorganized and sparse in a large 3D space. Classic Farthest Point Sampling (FPS) and KNN or ball query in PointNet++~\cite{qi2017pointnet++} are time-consuming \cite{rethage2018fully,liu2019point,hu2020randla,xu2020grid,hu2021learning}. Therefore, 
projection-aware set conv layer is proposed to sample center points and group neighborhood points effeciently. The proposed projection-aware set conv layer is shown in Fig. \ref{fig:new_ops1}, including stride-based sampling, projection-aware grouping, 3D distance-based filtering, and feature aggregation. The details are as follows: 

{\bf1) Stride-based Sampling (SBS).} The proposed stride-based sampling strategy is similar to the kernel center selecting of 2D convolution. The kernel centers are a series of fixed-position indexes according to the stride. Therefore, the indexes of sampling points can be directly generated when the sampling stride is set for a $H \times W \times 3$ XYZ coordinate map. Directly generating the indexes of the sampling points is more effective than random sampling. Meanwhile, the sampled points are more evenly distributed in the space.

{\bf2) Projection-aware Grouping.} Because of the line scan characteristics of LiDAR, points are usually closer in 3D space when they are closer in cylindrical projection space, except for the edges of the foreground and background. Therefore, similar to 2D convolution, for each sampled center point, we set a fixed-size kernel, and take the points in the kernel as the grouping points. {The grouping operation is similar to 3D-MiniNet \cite{alonso20203d}. However, the grouping will introduce distant points at the edges of the foreground and background. 3D-MiniNet \cite{alonso20203d} does not handle this, and 3D distance-based filtering is proposed in this paper to address this as follows.}

{\bf3) 3D Distance-based Filtering.}  {This 3D filtering operation is similar to the point cloud post-processing method of RangeNet++ \cite{milioto2019rangenet++}, but we integrate this process into the feature learning process for 3D point clouds. In addition, the filtering operation for \cite{milioto2019rangenet++} is after the KNN operation, while we need to keep the number of selected points fixed while filtering out unreasonable points for stable feature learning. Therefore, in our method, the points exceeding a fixed distance are firstly filtered out, and then points with a fixed number $K$ are randomly selected from the remaining points.} This method is also different from the ball query method \cite{qi2017pointnet++}. The  difference is that the prior information of the cylindrical projection helps filter many distant points in advance, which greatly speeds up the calculation as shown in subsequent experiments.
As can be seen from Fig. \ref{fig:new_ops1}, for feature extraction of a car, after 3D distance-based filtering, interfering points from the background are filtered out.  
Projection-aware grouping and 3D distance-based filtering are performed for each sampled point, and are calculated in parallel on the GPU.

\begin{figure}[t]
	\centering
	\resizebox{1.0\columnwidth}{!}
	{
		\includegraphics[scale=1.00]{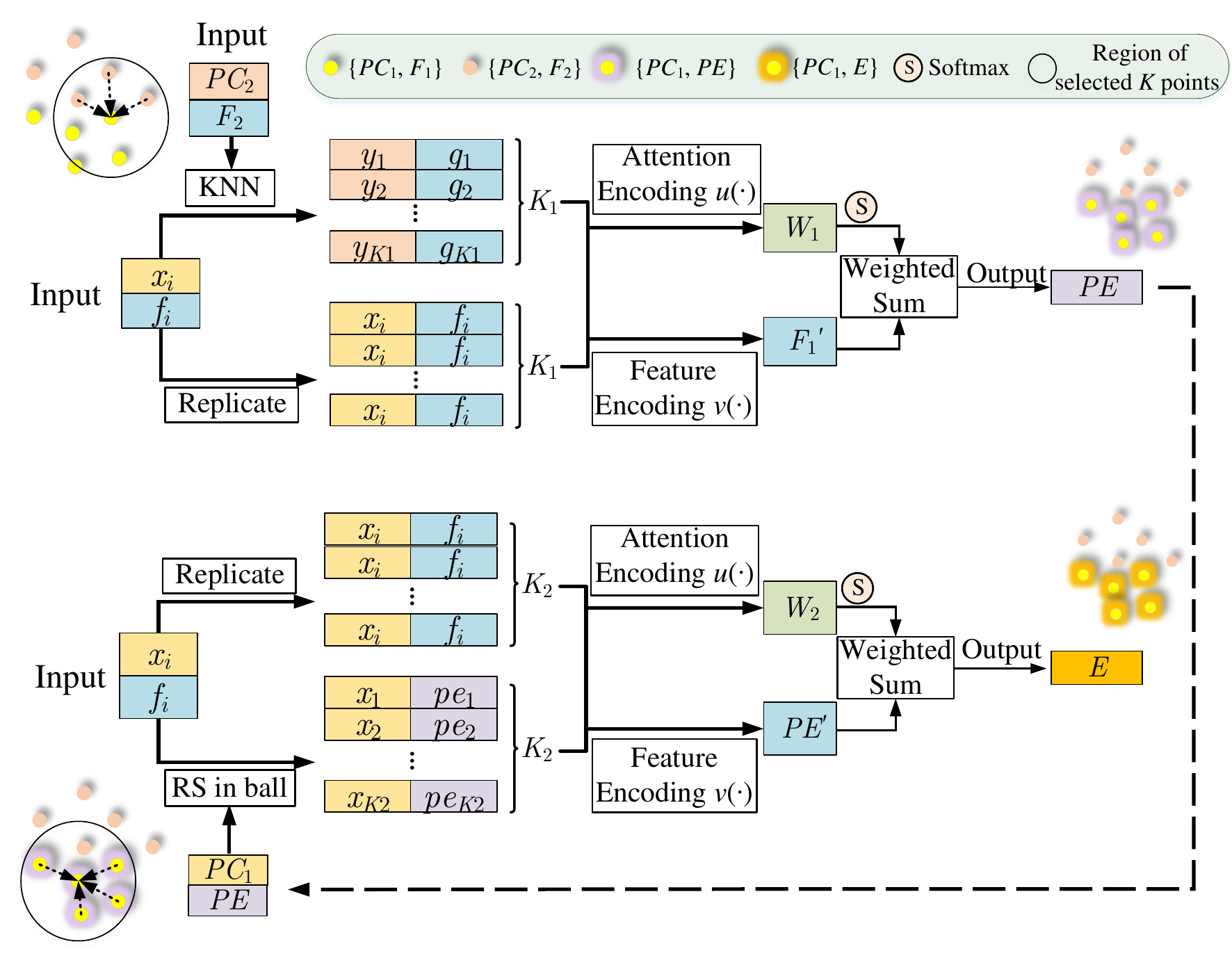}}
	\vspace{-6mm}
	\caption{Attention Cost-volume. This module takes two frames of point clouds with their local features as input and associates the two point clouds. Finally, the module outputs the embedding features located in $PC_1$.}
	\label{fig:cost}
\end{figure}

\begin{figure*}[t]
	\centering
	\resizebox{1.00\textwidth}{!}
	{
		\includegraphics[scale=1.00]{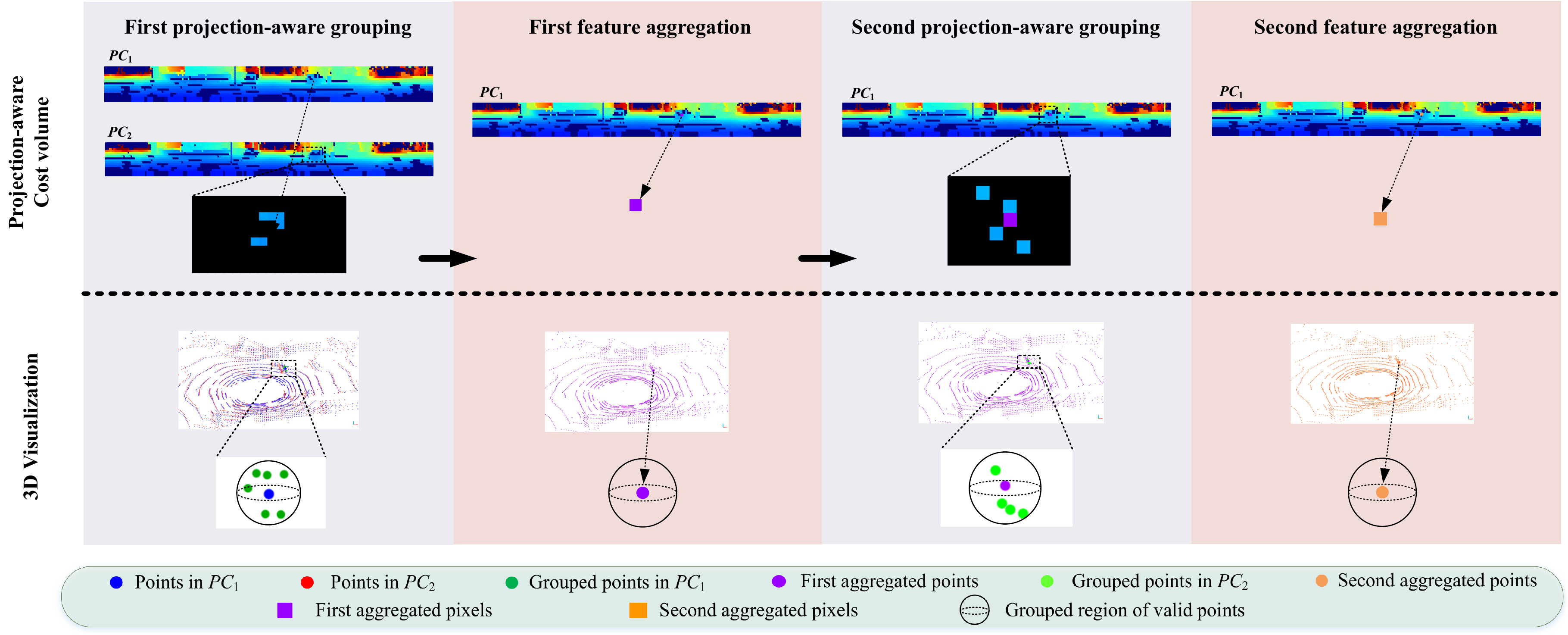}}
	\vspace{-6mm}
	\caption{The details of projection-aware cost volume modules. The module is composed of two projection-aware grouping parts, and two feature aggregation parts. The output is $PC_1$ with aggregated feature from $PC_2$ on the corresponding projected 2D plane. }
	\label{fig:new_ops2}
\end{figure*}

{\bf4) Feature Aggregation.} After getting the center point and the surrounding grouped points, Multi-Layer Perceptron (MLP) and max pooling are used for feature aggregation. The formula is:
\begin{equation} {f_i} = \mathop {MAX}\limits_{k = 1,2,...K} (MLP((x_i^k - x_i) \oplus f_i^k \oplus f_i^c))\label{eqn:feature_aggre},\end{equation}
where $x_i$ is the obtained $i$-th sampled point by stride-based sampling. And $K$ points $x_i^k$ $(k=1,2,...,K)$ are selected by Random Selecting (RS) around $x_i$ in a ball. $f_i^c$ and $f_i^k$ are the local features of $x_i$ and $x_i^k$ (they are null for the first layer in the pyramid). ${f_i}$ is the output feature located at the central point $x_i$. $\oplus$ denotes the concatenation of two vectors, and ${MAX}(\cdot)$ indicates the max pooling operation. A projection-aware siamese feature pyramid consisting of several projection-aware set conv layers is built to encode and extract the hierarchical features of each point cloud as shown in Fig.~\ref{fig:network}. The siamese pyramid \cite{chopra2005learning} means that the learning parameters of the built pyramid are shared for these two point clouds.

\subsection{Projection-aware Attentive Cost Volume}\label{sec:cost}

Next, inspired by \cite{wang2021hierarchical}, we introduce the projection-aware attentive cost volume to associate two point clouds. The cost volume generates point embedding features by associating two point clouds after the feature pyramid. 
The embedding features contain the point correlation information between two point clouds. The basic point cost volume is shown in Fig.~\ref{fig:cost}.  $F_1 = \{ {f_i}|{f_i} \in {\mathbb{R}^c}\} _{i = 1}^n$ are the features of point cloud $PC_1 = \{ {x_i}|{x_i} \in {\mathbb{R}^3}\} _{i = 1}^n$ and $F_2 = \{ {g_j}|{g_j} \in {\mathbb{R}^c}\} _{j = 1}^n$ are the features of point cloud $PC_2 = \{ {y_j}|{y_i} \in {\mathbb{R}^3}\} _{i = 1}^n$. First, $PC_2$ are grouped with each point $x_i$ of $PC_1$ as the center. This process finds the nearest $K_1$ neighbor points $y_j^k$ $(k=1,2...,K_1)$ in the 3D Euclidean space for each $x_i$. Then the feature correlation aggregation is performed based on attention to get the first attentive flow embedding $PE = \{pe_{i}|{pe_{i}\in \mathbb{R}^c}\}$ located at $PC_1$. After that, the first attentive flow embeddings $PE$ are grouped with each point $x_i$ as the center, and then the grouped first attentive flow embeddings $pe_{i}^k$ located at $x_i^k$ $(k=1,2...,K_2)$ are re-aggregated based on attention to obtain the final attentive flow embedding $e_{i}$. The calculation progress is as follows:
\begin{equation}w_{1, i}^k = softmax(u(x_i,y_j^k,f_i,g_j^k))_{k = 1}^{K_1},\end{equation}
\begin{equation}pe_{i} = \sum\limits_{k = 1}^{k_1} {{w_{1,i}^k} \odot v(x_i,y_j^k,f_i,g_j^k)},\end{equation}
\begin{equation}w_{2,i}^k = softmax(u(x_i,x_i^k,pe_{i},pe_{i}^k))_{k = 1}^{K_2},\end{equation}
\begin{equation}e_{i} = \sum\limits_{k = 1}^{k_2} {w_{2,i}^k \odot v(x_i,x_i^k,pe_{i},pe_{i}^k)},\end{equation}
where $g_j^k$ represents the local feature of $y_j^k$ queried in $PC_2$.  $\odot$ represents dot product. $u(\cdot)$ and $v(\cdot)$ represent the attention encoding and feature encoding functions referring to \cite{wang2021hierarchical}. The output $E = \{ {e_{i}}|{e_{i}} \in {\mathbb{R}^c}\} _{i = 1}^n$ is the final embeding features in $PC_1$.

\begin{figure}[t]
	\centering
	\vspace{-0mm}
	\resizebox{1.00\columnwidth}{!}
	{
		\includegraphics[scale=1.00]{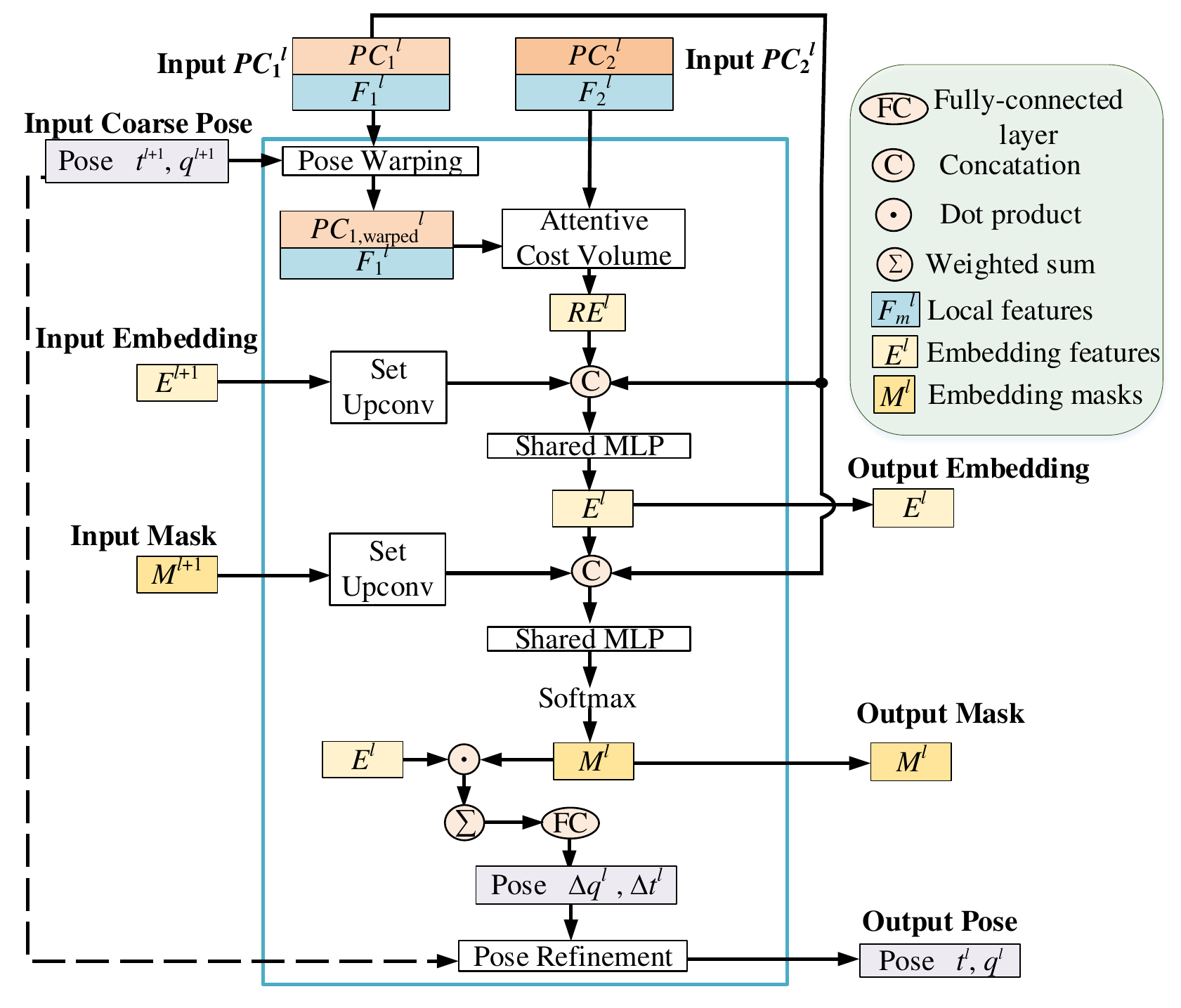}}
	\vspace{-7mm}
	\caption{The details of the proposed Pose Warp-Refinement module at the $l$-th level.}
	\label{fig:warping}
\end{figure}

\begin{figure*}[t]
	\centering
	\resizebox{1.00\textwidth}{!}
	{
		\includegraphics[scale=1.00]{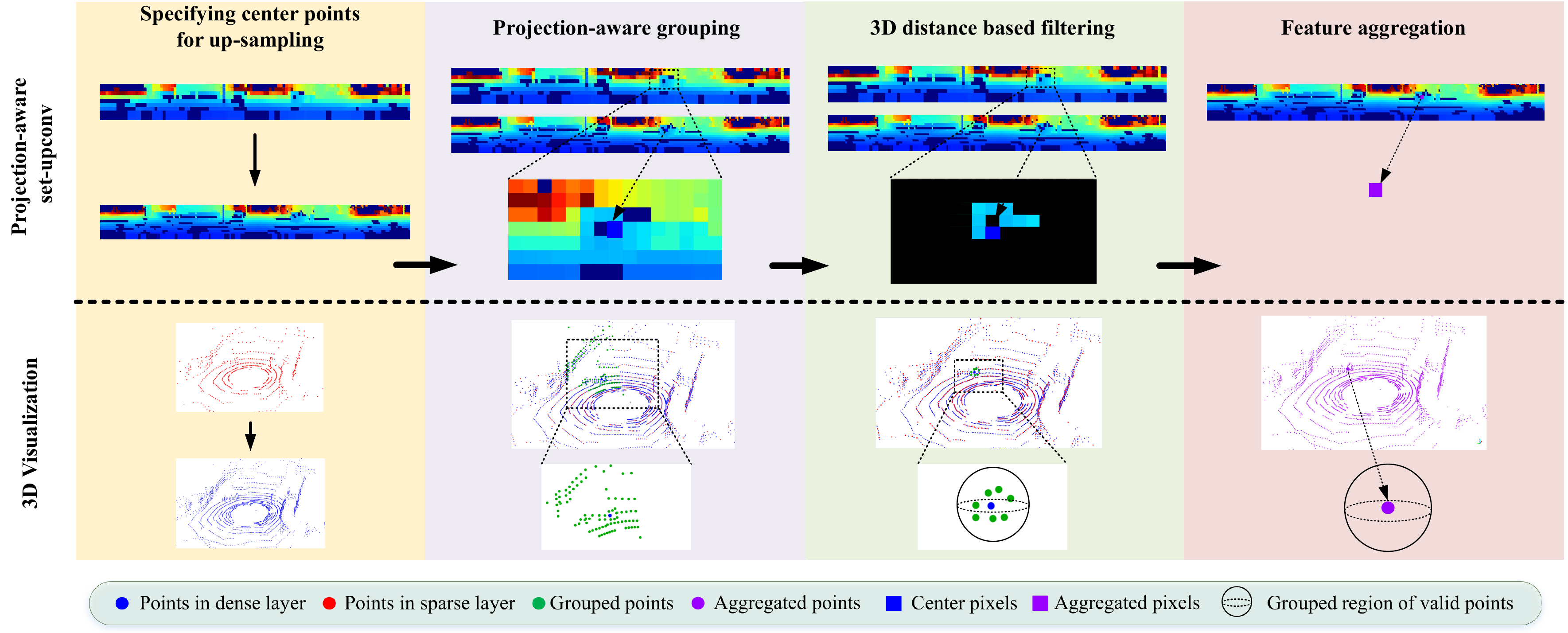}}
	\vspace{-6mm}
	\caption{The details of projection-aware set-upconv modules. The module is composed of four key parts: center points specification, projection-aware grouping, 3D distance-based filtering, and feature aggregation. The output is the points in dense layer with aggregated feature on their corresponding projected 2D plane.}
	\label{fig:new_ops3}
\end{figure*}

Due to the inefficient KNN operations, we restrict KNN to be proceeded in a fixed-size kernel on the projection plane, and refer the revised attentive cost volume as projection-aware attentive cost volume. The main improvement is to group neighborhood points in a kernel range on the 2D projection plane as shown in Fig. \ref{fig:new_ops2}. For the first grouping of $PC_2$ with $x_1$ as the center, the nearest neighbor points are queried in the 3D Euclidean space but the search range is restricted in a kernel on the projection plane of $PC_2$. For the second grouping of the first attentive flow embeddings $PE$, Random Selecting (RS) based neighbor query in a ball is performed while the grouping process is also restricted in a kernel range on the projection plane of $PC_1$, like projection-aware set conv in Sec. \ref{sec:pyramid}. These projection-aware grouping methods improve the efficiency.

\subsection{Hierarchical Embedding Mask Optimization}\label{sec:mask}
It is a new problem to convert the embedding features $E$ to a global consistent pose transformation between two frames. In this subsection, a novel embedding mask is proposed to generate pose transformation from embedding features.

It should be noted that some points may belong to dynamic objects or be occluded in the other frame. It is necessary to filter out these points and keep the points that are of value to the LiDAR odometry task. To deal with this, the embedding features $E = \{ {e_{i}}|{e_{i}} \in {\mathbb{R}^c}\} _{i = 1}^n$ and the features $F_1$ of $PC_1$ are input to a shared MLP followed by the softmax operation along the point dimension to obtain the embedding mask (as the initial embedding mask in Fig.~\ref{fig:network}):
\begin{equation}
	{M} = softmax(sharedMLP({E} \oplus F_1)),
	\end{equation}
where $M = \{ {m_{i}}|{m_{i}} \in {\mathbb{R}^c}\} _{i = 1}^n$ represents trainable masks for prioritizing embedding features of $n$ points in $PC_1$. Each point has a characteristic weight ranging from 0 to 1. The smaller the weight of a point is, the more likely the point needs to be filtered out, and vice versa. Then, the quaternion  $q \in {\mathbb{R}^4}$ and translation vector $t \in {\mathbb{R}^3}$ can be generated by weighting embedding features and FC layers separately, and $q$ is normalized to accord with the characteristic of rotation:
\begin{equation}
	q = \frac{{FC(\sum\limits_{i = 1}^n {{e_i} \odot {m_i}} )}}{{\left| {FC(\sum\limits_{i = 1}^n {{e_i} \odot {m_i}} )} \right|}}\label{eqn:q},\end{equation}
\vspace{-0.2cm}
\begin{equation}
	t = FC(\sum\limits_{i = 1}^n {{e_{i}} \odot {m_{i}}} )\label{eqn:t}.\end{equation}

The trainable mask $M$ is also part of the hierarchical refinement. As shown in Fig.~\ref{fig:network}, the embedding mask is propagated to denser layers of the point cloud just like embedding features $E$ and pose. The embedding mask is also optimized in a coarse-to-fine approach during the warp-refinement process, making the final mask estimation and the calculation of pose transformation accurate and reliable. We call this process the hierarchical embedding mask optimization.

\subsection{Iterative Pose Warp-Refinement}\label{sec:refine}
To achieve the coarse-to-fine refinement process in an end-to-end fashion, we propose the differentiable warp-refinement module based on pose transformation as shown in Fig.~\ref{fig:warping}. This module contains several key parts: projection-aware set upconv layer, projection-aware pose warping, embedding feature and embedding mask refinement, and pose refinement.

\vspace{5pt}
\noindent{}{\bf Projection-aware Set Upconv Layer:}
To refine the pose estimation in a coarse-to-fine approach, the projection-aware set upconv layer, inspired by \cite{liu2019flownet3d}, is proposed here to enable features of point cloud to propagate from sparse level to dense level. Similar to the projection-aware set conv, the grouping process for projection-aware set upconv is also constrained in a kernel on a 2D plane. As shown in Fig. \ref{fig:new_ops3}, there are also four steps, like projection-aware set conv. They are specifying the center point for up-sampling, projection-aware grouping, 3D distance-based filtering, and feature aggregration, respectively. First, the center points for up-sampling the sparse embedding features $E^{l+1}$ and embedding masks $M^{l+1}$ of $l+1$ layer are specified by the dense $PC_1^l$. Then, the center points are used to group neighborhood points in a fixed-size kernel like the projection-aware grouping in projection-aware set conv, but from the sparse $PC_1^{l+1}$. 3D distance-based filtering is used to filter the far points in the kernel. Finally, the same method as fomular (\ref{eqn:feature_aggre}) is used for the feature aggregration of the final grouped points to the dense $PC_1^l$.

Embedding features $E^{l+1}$ and embedding masks $M^{l+1}$ of $l+1$ layer are propagated through projection-aware set upconv layer to obtain coarse embedding features $CE^{l}=\{ {ce_{i}^l}|{ce_{i}^l} \in {\mathbb{R}^{c^l}}\} _{i = 1}^{n^l}$ and coarse embedding masks $CM^{l}= \{ {cm_{i}^l}|{cm_{i}^l} \in {\mathbb{R}^{c^l}}\} _{i = 1}^{n^l}$ that need to be optimized at the $l$-th level.

\setlength{\tabcolsep}{6.0mm}
\begin{table*}[t]
	\centering
	\footnotesize
	\caption{Detailed network parameters in EfficientLO-Net. Set conv layer uses the stride-based sampling to obtain the sampling points, and the sampling rate is less than $1$. For the set upconv layer, skip connections are used to propagate the sparse features to dense features, and the sampling rate is larger than $1$. $K$ points are selected by the $K$ Nearest Neighbors (KNN) or Random Selecting (RS) for the grouping process of set conv layer, set upconv layer, and attentive cost volume layer.  MLP width means the number of output channels for each layer of MLP. 
	}
	\vspace{-19pt}
	\begin{center}
		\resizebox{1.00\textwidth}{!}
		{
			\begin{tabular}{@{}cccccc@{}}
				\toprule
				Module & \multicolumn{2}{c}{Layer type}  & Sample rate &$K$& MLP width \\ \midrule
				&\multicolumn{2}{c}{Set conv layer for $PC_{1}^0$ and $PC_{2}^0$}  & 1/32& 32 & [8,8,16] \\
				&\multicolumn{2}{c}{Set conv layer for $PC_{1}^1$ and $PC_{2}^1$}  & 1/4& 32  & [16,16,32]\\
				&\multicolumn{2}{c}{Set conv layer for $PC_{1}^2$ and $PC_{2}^2$} & 1/4 &16& [32,32,64] \\
				\multirow{-4}{*}{\begin{tabular}[c]{@{}c@{}}Siamese Point \\ Feature Pyramid \end{tabular}}&\multicolumn{2}{c}{Set conv layer for $PC_{1}^3$ and $PC_{2}^3$} & 1/2 & 16 & [64,64,128]  \\
				
				\cline{1-6}\noalign{\smallskip}
				
				Attentive Cost Volume \\(on the penultimate level) 
				&\multicolumn{2}{c}{\multirow{-2}{*}{Attentive cost volume for $E_{coarse}$}}  & \multirow{-2}{*}1 &  \multirow{-2}{*}{4, 32}  & \multirow{-2}{*}{[128,64,64], [128,64]} \\
				
				\cline{1-6}\noalign{\smallskip}
				
				&\multicolumn{2}{c}{Set conv layer for $E^3$}  & 1/2 & 16 & [128,64,64]\\
				
				&\multicolumn{2}{c}{Shared MLP for $M^3$}     & 1 & ---  & [128,64] \\
				
				&\multicolumn{2}{c}{FC for $q^3$, FC for $t^3$}  & 1 & --- & [4], [3] \\
				
				\cline{1-6}\noalign{\smallskip}
				\multirow{-5.5}{*}{\begin{tabular}[c]{@{}c@{}}Generation of Initial \\Embedding Mask and Pose \end{tabular}}	
				&&Attentive cost volume for $RE^2$  & 1   & 4, 6 & [128,64,64], [128,64] \\
				&&Set upconv for $CE^2$  & 2& 8  & [128,64], [64] \\	
				&&Shared MLP for $E^2$  &  1 & ---  & [128,64] \\
				
				&&Set upconv  for $CM^2$     & 2  & 8    & [128,64], [64] \\
				&&Shared MLP  for $M^2$     &   1  & ---  & [128,64] \\
				
				&\multirow{-6}{*}{\begin{tabular}[c]{@{}c@{}}Pose Warp-Refinement \end{tabular}}  
				&FC for $q^2$, FC for $t^2$   & 1& --- & [4], [3] \\
				
				\cline{2-6}\noalign{\smallskip}					
				&&Attentive cost volume for $RE^1$  & 1   & 4, 6  & [128,64,64], [128,64] \\
				&&Set upconv for $CE^1$  & 4& 8  & [128,64], [64] \\	
				&&Shared MLP for $E^1$  &  1  & --- & [128,64] \\
				
				&&Set upconv  for $CM^1$     & 4 & 8    & [128,64], [64] \\
				&&Shared MLP  for $M^1$   &   1  & ---  & [128,64] \\
				
				&\multirow{-6}{*}{\begin{tabular}[c]{@{}c@{}}Pose Warp-Refinement \end{tabular}}  
				&FC for $q^1$, FC for $t^1$  & 1  & ---& [4], [3] \\
				
				\cline{2-6}\noalign{\smallskip}	
				
				&&Attentive cost volume for $RE^0$  & 1  & 4, 6  & [128,64,64], [128,64] \\
				&&Set upconv for $CE^0$ & 4& 8   & [128,64], [64] \\	
				&&Shared MLP for $E^0$   &  1 & --- & [128,64] \\
				
				&&Set upconv  for $CM^0$     & 4 & 8    & [128,64], [64] \\
				&&Shared MLP  for $M^0$     &   1  & ---  & [128,64] \\
				
				&\multirow{-6}{*}{\begin{tabular}[c]{@{}c@{}}Pose Warp-Refinement \end{tabular}}  
				&FC for $q^0$, FC for $t^0$  & 1  & ---& [4], [3] \\
				
				\bottomrule
				\multirow{-20.5}{*}{\begin{tabular}[c]{@{}c@{}}Iterative Pose\\Warp-Refinement \end{tabular}} 
		\end{tabular}	}
	\end{center}
	\vspace{-2pt}
	\label{table:parameters}
\end{table*}

\vspace{5pt}
\noindent{}{\bf Projection-aware Pose Warping:}
The process of pose warping means that the quaternion $q^{l+1}$ and translation vector $t^{l+1}$ from the ${(l+1)}$-th level are applied to warp $PC_1^l=\{ {x_{i}^l}|{x_{i}^l} \in {\mathbb{R}^{c^l}}\} _{i = 1}^{n^l}$ to generate $PC_{1,warped}^l=\{ {x_{i,warped}^l}|{x_{i,warped}^l} \in {\mathbb{R}^{c^l}}\} _{i = 1}^{n^l}$. The warped $PC_{1,warped}^l$ is closer to $PC_2^l$ than original $PC_{1}^l$, which makes the residual motion estimation easier at the $l$-th level. The equation of pose warping is as follows:
\begin{equation}
	[0,x_{i,warped}^l] = {q^{l+1}}[0,x_{i}^l](q^{l+1})^{-1}+[0, t^{l+1}].\end{equation}

Then, the warped $PC_{1,warped}^l$ is projected again as in Sec. \ref{sec:Representation} to get a new XYZ coordinate map closer to $PC_2^l$.
 The attentive cost volume between $PC{_{1,warped}^l}$ and $PC_2^l$ is re-calculated to estimate the residual motion. Following the approach introduced in Sec.~\ref{sec:cost}, re-embedding features between $PC{_{1,warped}^l}$ and $PC_2^l$ are calculated and denoted as $RE^l = \{ {re_{i}^l}|{re_{i}^l} \in {\mathbb{R}^{c^l}}\} _{i = 1}^{n^l}$.

\vspace{5pt}
\noindent{}{\bf Embedding Feature and Embedding Mask Refinement:}
The generated coarse embedding feature $ce_i^{l}$, the re-embedding feature $re_{i}^l$, and the features $f_i^l$ of $PC_1^l$ are concatenated and input to a shared MLP to obtain embedding features $E^{l}=\{ {e_{i}^l}|{e_{i}^l} \in {\mathbb{R}^{c^l}}\} _{i = 1}^{n^l}$ at the $l$-th level:
\begin{equation}
	\vspace{-0.1cm}
	e_{i}^l = MLP(ce_{i}^l \oplus re_{i}^{l}  \oplus f_i^l).
	\end{equation}
The output of this MLP is the optimized embedding features of $l$-th level, which will not only participate in the following pose generation operation but also be output as the input to the next level warp-refinement module. 

Like the refinement of the embedding feature, the newly generated embedding feature $e_{i}^l$, the generated coarse embedding mask $cm_i^{l}$, and the local feature $f_i^l$ of $PC_1^l$ are concatenated and input to a shared MLP and softmax operation along the point dimension to obtain the embedding mask $M^{l}= \{ {m_{i}^l}|{m_{i}^l} \in {\mathbb{R}^{c^l}}\} _{i = 1}^{n^l}$ at the $l$-th level:
\begin{equation}
	M^l = softmax(sharedMLP(E^l \oplus CM^{l}  \oplus F^l)).
	\vspace{-5pt}
\end{equation}

\vspace{5pt}
\noindent{}{\bf Pose Refinement:}
The residual $\Delta q^l$ and $\Delta t^l$ can be obtained from the refined embedding features and mask following the formulas~(\ref{eqn:q}) and (\ref{eqn:t}) in Sec.~\ref{sec:mask}. Lastly, the refined quaternion $q^l$ and translation vector $t^l$ of the $l$-th level can be calculated by:
\begin{equation}{q^l} = {\Delta q^{l}}{q^{l + 1}},\end{equation}
\begin{equation}[0,t^l] = {\Delta q^{l}}[0,t^{l+1}](\Delta q^{l})^{-1}+[0, \Delta t^{l}].
\end{equation}

As shown in Fig. \ref{fig:projection}, the iterative pose warp-refinement part of the network uses the pose warp-refinement module three times.
There is an projection-aware set upconv operation for each pose refinement to make the point cloud denser and finer. Therefore, coarse inter-frame point cloud correlation and initial pose estimation are performed at a level with fewer points, which reduces calculation. The refinement process is carried out at a level that contains more points, which brings more details to get a finer registration. The structure of PWC allows our LiDAR odometry method to obtain higher accuracy while using as few calculations as possible.

\subsection{Training Loss}
The network outputs quaternion $q^l$ and translation vector $t^l$ from four different levels of point clouds. The outputs of each level will enter into a designed loss function and be used to calculate the supervised loss $\ell^l$. Due to the different scales and units between translation vector $t$ and quaternion $q$, two learnable parameters $s_x$ and $s_q$ are introduced like previous deep odometry work \cite{li2019net}. The training loss function at the $l$-th level is: \begin{equation}\begin{gathered}
		{\ell ^l} = {\left\| {{t_{gt}} - {t^l}} \right\|}exp( - {s_x}) + {s_x} \hfill \\
		{\text{   }} + {\left\| {{q_{gt}} - \frac{{{q^l}}}{{\left\| {{q^l}} \right\|}}} \right\|_2}exp( - {s_q}) + {s_q},  
	\end{gathered} \label{eqn:sxsq} \end{equation}
where $\left\|  \cdot  \right\|$ and ${\left\|  \cdot  \right\|_2}$ denote the ${\ell _1}$-norm and the ${\ell _2}$-norm respectively. ${{t_{gt}}}$ and ${{q_{gt}}}$ are the ground-truth translation vector and quaternion respectively generated by the ground-truth pose transformation matrix. Then, a multi-scale supervised approach is adopted. The total training loss $\ell$ is:
\begin{equation}\ell  = \sum\nolimits_{l = 1}^L {{\alpha ^l}} {\ell ^l}\label{eqn:l},\end{equation}
where $L$ is the total number of warp-refinement levels and ${{\alpha ^l}}$ is a hyperparameter denoting the weight of the $l$-th level.

\section{Implementation}\label{sec:implementation}
\subsection{Dataset Processing}
KITTI odometry dataset \cite{geiger2012we,geiger2013vision} is composed of 22 independent sequences. The Velodyne LiDAR point clouds in the dataset are used in our experiments. All scans of point clouds have XYZ coordinates and the reflectance information. Sequences 00-10 (23201 scans) contain ground truth pose (trajectory), while there is no ground truth publicly available for the remaining sequences 11-21 (20351 scans). By driving under different road environments, such as highways, residential roads, and campus roads, the sampling vehicle captures point clouds for the LiDAR odometry task from different environments.

\vspace{5pt}
\noindent{}{\bf Data Preprocessing:}
Only coordinates of LiDAR points are used in our method. As described in Sec. \ref{sec:Representation}, cylindrical projection is used to project and organize 3D point clouds to obtain projection-aware representation of point clouds to speed up subsequent projection-aware operations.
Moreover, the point cloud collected by the LiDAR sensor often contains outliers at the edge of the point cloud in each frame. This is often because objects are far away from the LiDAR sensor, thus forming incomplete point clouds in the edge. To filter out these outlier points, the points out of the $30 \times 30m^2$ square area around the vehicle are filtered out for each point cloud.

\vspace{5pt}
\noindent{}{\bf Data Augmentation:}
We augment the training dataset by the augmentation matrix ${T_{aug}}$, generated by the rotation matrix ${R_{aug}}$ and translation vector ${t_{aug}}$.
Varied values of yaw-pitch-roll Euler angles are generated by Gaussian distribution around $0^{\circ}$. Then the ${R_{aug}}$ can be obtained from these random Euler angles. Similarly, the ${t_{aug}}$ is generated by the same process.
The composed ${T_{aug}}$ is then used to augment the $PC_1$ to obtain new point clouds $PC_{1,aug}$ by:
\begin{equation} 
	{PC_{1,aug}} = {T_{aug}}{PC_1}.\end{equation}
Correspondingly, the ground truth transformation matrix is also modified as:
\begin{equation}
	{T_{trans}} = {T_{aug}}{T_p},\end{equation}
where ${T_p}$ denotes the original ground truth pose transformation matrix from $PC_1$ to $PC_2$.
${T_{trans}}$ is then used to generate ${q_{gt}}$ and ${t_{gt}}$ to supervise the training of the network.

\subsection{Parameters}

During training and evaluation, the network input is $H \times W \times 3$ projection-aware representation of 3D point cloud. In the experiment, $H$ is set as 64 and $W$ is set as 1800, according to the characteristics of Velodyne LiDAR in KITTI dataset. Each layer in MLP contains the ReLU activation function, except for the FC layer. For shared MLP, $1 \times 1$ convolution with $1$ stride is the implement manner. The detailed layer parameters including the sample rate of each sampling layer, $K$ values in each grouping layer, and each linear layer width in MLP are described in Table~\ref{table:parameters}. All training and evaluation experiments are conducted on a single NVIDIA Titan RTX GPU with TensorFlow 1.9.0. The Adam optimizer is adopted with ${\beta _1} = 0.9$, ${\beta _2} = 0.999$. The initial learning rate is 0.001 and exponentially decays every 200000 steps until 0.00001. The initial values of the trainable parameters $s_x$ and $s_q$ are set as 0.0 and -2.5 respectively in formula~(\ref{eqn:sxsq}). For formula~(\ref{eqn:l}), ${\alpha _1}$ = 1.6, ${\alpha _2}$ = 0.8, ${\alpha _3}$ = 0.4, and $L=4$. The batch size is 8. 

 \setlength{\tabcolsep}{0.9mm}
 \begin{table*}[t]
 	\centering
 	\footnotesize
 	\caption{The LiDAR odometry experiment results on KITTI odometry dataset \cite{geiger2013vision}. $t_{rel}$ and $r_{rel}$ mean the average translational RMSE (\%) and rotational RMSE ($^{\circ}$/100m) respectively on all possible subsequences in the length of $100,200,...,800\,m$. `$^*$' means the training sequence. LOAM \cite{zhang2017low}  is a complete SLAM system, including back-end optimization and others only include odometry. A-LOAM \cite{aloam} is an advanced re-implementation of LOAM, and improves the basic codes of LOAM. The results of Full A-LOAM and A-LOAM w/o mapping are obtained by running their published codes. The best results are bold.}
 	\vspace{-20pt}	
 	\begin{center}
 		\resizebox{1.0\textwidth}{!}
 		{
 			\begin{tabular}{l||cc|cc|cc|cc|cc|cc|cc|cc|cc|cc|cc||cc}
 				\toprule
 				&  \multicolumn{2}{c|}{00$^*$}  &\multicolumn{2}{c|}{01$^*$}      & \multicolumn{2}{c|}{02$^*$} & \multicolumn{2}{c|}{03$^*$} &  \multicolumn{2}{c|}{04$^*$} & \multicolumn{2}{c|}{05$^*$} & \multicolumn{2}{c|}{06$^*$} & \multicolumn{2}{c|}{07} & \multicolumn{2}{c|}{08} & \multicolumn{2}{c|}{09} &\multicolumn{2}{c||}{10} &\multicolumn{2}{c}{Mean on 07-10} \\ 
 				\cline{2-25}\noalign{\smallskip}
 				
 				\multirow{-2}{*}{\begin{tabular}[c]{@{}c@{}}Method \end{tabular}}
 				&  $t_{rel}$  & $r_{rel}$   & $t_{rel}$                       & $r_{rel}$               & $t_{rel}$                          & $r_{rel}$   & $t_{rel}$ & $r_{rel}$   & $t_{rel}$                          & $r_{rel}$   & $t_{rel}$ & $r_{rel}$    & $t_{rel}$                          & $r_{rel}$   & $t_{rel}$ & $r_{rel}$ & $t_{rel}$                          & $r_{rel}$   & $t_{rel}$ & $r_{rel}$    & $t_{rel}$ & $r_{rel}$  & $t_{rel}$ & $r_{rel}$       \\
 				\hline\hline
 				\noalign{\smallskip}
 				
 				Full LOAM \cite{zhang2017low}    
 				&0.78 &	0.53 
 				&1.43 &	0.55 
 				&0.92 & 0.55 
 				&0.86 &	0.65 
 				&0.71 & 0.50 
 				&0.57 &	0.38 
 				&0.65 & 0.39 
 				&0.63 & 0.50 
 				&1.12 & 0.44 
 				&0.77 & 0.48 
 				&\bf0.79 & 0.57 
 				& 0.828 &  0.498
 				\\
 				Full A-LOAM \cite{aloam}   
 				&\bf0.76 &	\bf0.31 
 				&1.97 &	0.50 
 				&4.53 & 1.45 
 				&0.93 &	0.49 
 				&0.62 & 0.39 
 				&0.48 &	0.25 
 				&0.61 & 0.28 
 				&\bf0.43 & \bf0.26 
 				&\bf1.06 &\bf0.32
 				&\bf0.73 &\bf0.31 
 				&1.02 &\bf0.40 
 				& 0.810 &\bf0.323
 				\\
 				\hline 
 				ICP-po2po    
 				&6.88&	2.99
 				&11.21&	2.58
 				& 8.21	&3.39
 				&11.07&	5.05
 				& 6.64	&4.02
 				& 3.97&	1.93
 				&1.95	&1.59
 				&5.17	&3.35
 				&10.04	&4.93
 				&6.93	&2.89
 				&8.91	&4.74
 				&7.763	&3.978
 				\\ 
 				
 				ICP-po2pl     
 				&3.80 &	1.73
 				&13.53 & 2.58 
 				&9.00 & 2.74 
 				&2.72 & 1.63 
 				&2.96 & 2.58 
 				&2.29 & 1.08 
 				&1.77 & 1.00 
 				&1.55 & 1.42 
 				&4.42 & 2.14 
 				&3.95 & 1.71 
 				&6.13 & 2.60 
 				&4.013 & 1.968
 				\\

 				GICP \cite{segal2009generalized}    
 				&1.29 & 0.64 
 				&4.39 & 0.91 
 				&2.53 & 0.77 
 				&1.68 & 1.08 
 				&3.76 & 1.07 
 				&1.02 & 0.54 
 				&0.92 & 0.46 
 				&0.64 & 0.45 
 				&1.58 & 0.75 
 				&1.97 & 0.77 
 				&1.31 &0.62 
 				&1.375 & 0.648
 				\\ 
 				CLS \cite{velas2016collar}    
 				&2.11 & 0.95 
 				&4.22 & 1.05 
 				&2.29 & 0.86
 				&1.63 & 1.09 
 				&1.59 & 0.71 
 				&1.98 & 0.92 
 				&0.92 & 0.46 
 				&1.04 & 0.73 
 				&2.14 & 1.05 
 				&1.95 & 0.92 
 				&3.46 & 1.28 
 				&2.148 &  0.995
 				\\ 
 				
 				Velas et al. \cite{velas2018cnn}    
 				&3.02 &	NA
 				&4.44 &	NA
 				&3.42 & NA
 				&4.94 &	NA
 				&1.77 & NA
 				&2.35 &	NA
 				&1.88 & NA
 				&1.77 & NA
 				&2.89 & NA
 				&4.94 & NA
 				&3.27 & NA
 				&3.218 & NA
 				
 				\\ 
 				LO-Net \cite{li2019net}    
 				&1.47 & 0.72 
 				&1.36 & 0.47 
 				&1.52 & 0.71 
 				&1.03 & 0.66 
 				&0.51 & 0.65 
 				&1.04 & 0.69 
 				&0.71 & 0.50 
 				&1.70 & 0.89 
 				&2.12 & 0.77 
 				&1.37 & 0.58 
 				&1.80 & 0.93 
 				&1.748 & 0.793
 				\\ 
 				DMLO \cite{li2020dmlo}    
 				& NA   & NA
 				& NA   & NA
 				& NA   & NA
 				& NA   & NA
 				& NA   & NA
 				& NA   & NA
 				& NA   & NA
 				&0.73 & 0.48
 				&1.08 & 0.42
 				&1.10 & 0.61
 				& 1.12 & 0.64
 				&1.008 & 0.538
 				\\
 				PWCLO-Net \cite{wang2021pwclo}    
				&0.78 &0.42
				&0.67 &0.23
				&0.86 & 0.41
				&0.76 & 0.44
				&0.37 & 0.40
				&0.45 & 0.27
				&\bf0.27 &\bf0.22
				&0.60 & 0.44
				&1.26 & 0.55
				&0.79 & 0.35
				&1.69 &0.62
				&1.085  & 0.490  
				\\
 				A-LOAM w/o mapping   
 				&4.08	&1.69	&3.31	&0.92	&7.33	&2.51	&4.31	&2.11	&1.60	&1.13	&4.09	&1.68	&1.03	&0.52	&2.89	&1.8	&4.82	&2.08	&5.76	&1.85	&3.61	&1.76	&3.894	&1.641
 				\\             					
 				Ours      
 				&0.83 &0.33
 				&\bf0.55 & \bf0.21
 				&\bf0.71 & \bf0.25
 				&\bf0.49 & \bf0.38
 				&\bf0.22 & \bf0.11
 				&\bf0.34 & \bf0.21
 				&0.36 &0.24
 				&0.46 &0.38
 				&1.14    &0.41
 				&0.78 &0.33
 				&0.80 &0.46
 				&\bf0.795  &0.395    
 				\\ \bottomrule
 			\end{tabular}
 		}
 	\end{center}
 	\vspace{-3pt}
 	\label{table:lidar}		
 \end{table*}
 
 \setlength{\tabcolsep}{0.9mm}
 \begin{table*}[t]
 	\footnotesize
 	\caption{The LiDAR odometry experiment results on KITTI odometry dataset \cite{geiger2013vision} compared with \cite{wang2021hierarchical}. HALFlow \cite{wang2021hierarchical} is trained on Flything3D scene flow dataset. HALFlow \cite{wang2021hierarchical}-Refine is trained on KITTI seq. 00-06 again after being trained on Flything3D dataset. This refined-training process generates the ground truth 3D scene flow from ground truth pose transformation based on the assumption that all points are static in the frame. Ours are only trained on KITTI seq. 00-06.}
 	\vspace{-14pt}
 	\begin{center}
 		\resizebox{0.99\textwidth}{!}
 		{
 			\begin{tabular}{l||cc|cc|cc|cc|cc|cc|cc|cc|cc|cc|cc||cc}
 				\toprule
 				&  \multicolumn{2}{c|}{00$^*$}  &\multicolumn{2}{c|}{01$^*$}      & \multicolumn{2}{c|}{02$^*$} & \multicolumn{2}{c|}{03$^*$} &  \multicolumn{2}{c|}{04$^*$} & \multicolumn{2}{c|}{05$^*$} & \multicolumn{2}{c|}{06$^*$} & \multicolumn{2}{c|}{07} & \multicolumn{2}{c|}{08} & \multicolumn{2}{c|}{09} &\multicolumn{2}{c||}{10} &\multicolumn{2}{c}{Mean on 07-10} \\ 
 				\cline{2-25}\noalign{\smallskip}
 				
 				\multirow{-2}{*}{\begin{tabular}[c]{@{}c@{}}Method \end{tabular}}
 				&  $t_{rel}$  & $r_{rel}$   & $t_{rel}$                       & $r_{rel}$               & $t_{rel}$                          & $r_{rel}$   & $t_{rel}$ & $r_{rel}$   & $t_{rel}$                          & $r_{rel}$   & $t_{rel}$ & $r_{rel}$    & $t_{rel}$                          & $r_{rel}$   & $t_{rel}$ & $r_{rel}$ & $t_{rel}$                          & $r_{rel}$   & $t_{rel}$ & $r_{rel}$    & $t_{rel}$ & $r_{rel}$  & $t_{rel}$ & $r_{rel}$       \\
 				
 				\hline\hline
 				\noalign{\smallskip}
 				
 				HALFlow \cite{wang2021hierarchical} 
 				&10.46&4.46
 				&72.21&14.66
 				&24.54&8.64
 				&15.21&8.03
 				&54.30&34.02
 				&10.55&4.12
 				&14.35&5.87   		
 				&14.24  &8.15   		
 				&24.59  & 9.64
 				&21.43 &  8.51
 				&19.03 &8.29
 				& 19.823 &8.648
 				\\ 
 				HALFlow \cite{wang2021hierarchical}-Refine 
 				&2.89  &1.26
 				&2.37  &0.72
 				&2.64  &1.12
 				&2.79  &2.12
 				&1.74  &0.88
 				&3.01  &1.29
 				&3.23  &1.14 		
 				&2.92  &1.82 		
 				&4.13  &1.60 
 				&3.05  &1.11 
 				& 3.62 &1.78
 				&3.430  &1.578
 				\\ 
 				Ours      
 				&\bf0.83 & \bf0.33
 				&\bf0.55 & \bf0.21
 				&\bf0.71 & \bf0.25
 				&\bf0.49 & \bf0.38
 				&\bf0.22 & \bf0.11
 				&\bf0.34 & \bf0.21
 				&\bf0.36 & \bf0.24
 				&\bf0.46 & \bf0.38
 				&\bf1.14    &\bf0.41
 				&\bf0.78 & \bf0.33
 				&\bf0.80 & \bf0.46
 				&\bf0.795  &\bf 0.395    
 				\\ \bottomrule
 			\end{tabular}
 		}
 	\end{center}
 	\vspace{-2pt}			
 	\label{table:wang}
 \end{table*}

 \setlength{\tabcolsep}{1.2mm}
 \begin{table*}[t]
 	\centering
 	\footnotesize
 	\caption{The LiDAR odometry experiment results on KITTI odometry dataset \cite{geiger2013vision} compared with \cite{zheng2020lodonet}. As \cite{zheng2020lodonet} is trained on sequences 00-06, 09-10  and tested on sequences 07-08, we train and test our model like this to make comparisons with \cite{zheng2020lodonet}.}
 	\vspace{-14pt}
 	\begin{center}
 		\resizebox{0.94\textwidth}{!}
 		{
 			\begin{tabular}{l||cc|cc|cc|cc|cc|cc|cc|cc|cc|cc|cc||cc}
 				\toprule
 				&  \multicolumn{2}{c|}{00$^*$}  &\multicolumn{2}{c|}{01$^*$}      & \multicolumn{2}{c|}{02$^*$} & \multicolumn{2}{c|}{03$^*$} &  \multicolumn{2}{c|}{04$^*$} & \multicolumn{2}{c|}{05$^*$} & \multicolumn{2}{c|}{06$^*$} & \multicolumn{2}{c|}{07} & \multicolumn{2}{c|}{08} & \multicolumn{2}{c|}{09$^*$} &\multicolumn{2}{c||}{10$^*$} &\multicolumn{2}{c}{Mean on 07-08} \\ 
 				\cline{2-25}\noalign{\smallskip}
 				
 				\multirow{-2}{*}{\begin{tabular}[c]{@{}c@{}}Method \end{tabular}}
 				&  $t_{rel}$  & $r_{rel}$   & $t_{rel}$                       & $r_{rel}$               & $t_{rel}$                          & $r_{rel}$   & $t_{rel}$ & $r_{rel}$   & $t_{rel}$                          & $r_{rel}$   & $t_{rel}$ & $r_{rel}$    & $t_{rel}$                          & $r_{rel}$   & $t_{rel}$ & $r_{rel}$ & $t_{rel}$                          & $r_{rel}$   & $t_{rel}$ & $r_{rel}$    & $t_{rel}$ & $r_{rel}$  & $t_{rel}$ & $r_{rel}$       \\
 				
 				\hline\hline
 				\noalign{\smallskip}
 				
 				LodoNet \cite{zheng2020lodonet}      
 				&1.43	&0.69
 				&0.96	&0.28
 				&1.46	&0.57
 				&2.12	&0.98
 				&0.65	&0.45
 				&1.07	&0.59
 				&0.62	&0.34
 				&1.86	&1.64
 				&2.04	&0.97
 				&0.63	&0.35
 				&1.18	&0.45
 				&1.950	&1.305		
 				\\                		
 				Ours      
 				&\bf0.57 & \bf0.26
 				&\bf0.53 & \bf0.20
 				&\bf0.60 & \bf0.23
 				&\bf0.41 & \bf0.29
 				&\bf0.20 & \bf0.13
 				&\bf0.46 & \bf0.25
 				&\bf0.28 & \bf0.18
 				&\bf0.60 & \bf0.45
 				&\bf1.15 & \bf0.46
 				&\bf0.50 & \bf0.20
 				&\bf0.70 &\bf0.38
 				&\bf0.875 &\bf0.455
 				\\ \bottomrule
 			\end{tabular}
 		}
 	\end{center}
 \vspace{-2pt}
 	\label{table:lidar_7_8}
 \end{table*}
 
 \setlength{\tabcolsep}{1.3mm}
 \begin{table}[t]
 	\centering
 	\footnotesize
 	\caption{The LiDAR odometry results on sequences 04 and 10 of KITTI odometry dataset \cite{geiger2013vision}. As \cite{wang2019deeppco} is trained on sequences 00-03, 05-09 and only reports testing results on sequences 04 and 10, we train and test our model like this to make comparisons with it.}	
 	\vspace{-12pt}
 	\begin{center}
 		\resizebox{0.8\columnwidth}{!}
 		{
 			\begin{tabular}{l||cc|cc||cc}
 				\toprule
 				&    \multicolumn{2}{c|}{04} &\multicolumn{2}{c||}{10} &\multicolumn{2}{c}{Mean} \\ 
 				\cline{2-7}\noalign{\smallskip}
 				
 				\multirow{-2}{*}{\begin{tabular}[c]{@{}c@{}}Method \end{tabular}}
 				&  $t_{rel}$  & $r_{rel}$   & $t_{rel}$                       & $r_{rel}$               & $t_{rel}$                          & $r_{rel}$      \\
 				\hline\hline
 				\noalign{\smallskip}

 				DeepPCO \cite{wang2019deeppco}    
 				
 				&2.63 & 3.05 
 				
 				&2.47 & 6.59 
 				&2.550 & 4.820
 				\\

 				Ours      
 				
 				&\bf0.59 & \bf0.37
 				
 				&\bf1.05 & \bf0.52
 				&\bf0.820    & \bf 0.445 
 				\\ \bottomrule
 			\end{tabular}
 		}
 	\end{center}
 	\vspace{-4pt}
 	\label{table:lidar4_10}
 \end{table}

\setlength{\tabcolsep}{1.3mm}
\begin{table}[t]
	\centering
	\footnotesize
	\caption{The LiDAR odometry results on sequences 09 and 10 of KITTI odometry dataset \cite{geiger2013vision}. As \cite{cho2020} applies unsupervised training on sequences 00-08, and reports testing results on sequences 09 and 10, we train and test our model like this to make fair comparisons.}	
	\vspace{-12pt}	
	\begin{center}
		\resizebox{0.8\columnwidth}{!}
		{
			\begin{tabular}{l||cc|cc||cc}
				\toprule
				&    \multicolumn{2}{c|}{09} &\multicolumn{2}{c||}{10} &\multicolumn{2}{c}{Mean} \\ 
				\cline{2-7}\noalign{\smallskip}
				
				\multirow{-2}{*}{\begin{tabular}[c]{@{}c@{}}Method \end{tabular}}
				&  $t_{rel}$  & $r_{rel}$   & $t_{rel}$                       & $r_{rel}$               & $t_{rel}$                          & $r_{rel}$      \\
				\hline\hline
				\noalign{\smallskip}

				Cho et al. \cite{cho2020}    
				
				&4.87 & 1.95 
				
				&5.02 & 1.83 
				&4.945 & 1.890
				\\

				Ours      
				
				&\bf0.90 & \bf0.37
				
				&\bf0.90 & \bf0.56
				&\bf0.900    & \bf 0.465 
				\\ \bottomrule
			\end{tabular}
		}
	\end{center}
	\vspace{-4pt}
	\label{table:lidar09_10}
\end{table}

For the data augmentation, the set standard deviation is different for each angle and each direction of the translation vector because of the motion characteristics of the car. We set the standard deviations as $0.05^{\circ}$, $0.01^{\circ}$, and $0.01^{\circ}$ for the yaw-pitch-roll Euler angles respectively. We set the standard deviations as $0.5m$, $0.1m$, and $0.05m$ for the XYZ of the translation respectively. In these Gaussian distributions, we select the data in the range of 2 times standard deviation around the mean value for data augmentation.

\section{Experimental Results}
In this section, the quantitative and qualitative results of the network performance on the LiDAR odometry task are demonstrated and compared with state-of-the-art methods. Then, the ablation studies are presented. Finally, the embedding mask and registration effect are visualized and discussed. 

\setlength{\tabcolsep}{1.3mm}
\begin{table}[t]
	\centering
	\footnotesize
	\caption{The LiDAR odometry results on official test sequences 11-21 of KITTI odometry dataset \cite{geiger2013vision}. We also tested the results of the four geometry-based methods on sequences 07-10 for comparison.}	
	\vspace{-12pt}	
	\begin{center}
		\resizebox{0.85\columnwidth}{!}
		{
			\begin{tabular}{l||cc||cc}
				\toprule
				&  \multicolumn{2}{c||}{Mean on 07-10} &\multicolumn{2}{c}{Mean on 11-21} \\ 
				\cline{2-5}\noalign{\smallskip}
				
				\multirow{-2}{*}{\begin{tabular}[c]{@{}c@{}}Method \end{tabular}}
				&  $t_{rel}$  & $r_{rel}$   & $t_{rel}$                       & $r_{rel}$                     \\
				\hline\hline
				\noalign{\smallskip}

				pyLiDAR (w/o mapping)&	1.603&	0.765&	2.91	&0.80\\
               LeGO-LOAM (w/o mapping)	&8.425	&4.730&	19.57	&4.93\\
               A-LOAM (w/o mapping)&	4.270&	1.873&	9.37	&1.92\\
               SuMa (w/o mapping)	&2.003	&0.970&	4.91	&1.53\\
               Ours&	\bf0.795	&\bf0.395	&\bf1.92	&\bf0.52
				\\ \bottomrule
			\end{tabular}
		}
	\end{center}
	\vspace{-4pt}
	\label{table:11-21}
\end{table}

\setlength{\tabcolsep}{1.5mm}
\begin{table*}[h]
	\footnotesize
	\centering
	\caption{The runtime comparison between ours and LO-Net. We report the average runtime of our models with one batch size on KITTI Seq. 00 for comparison with LO-Net.}
	\vspace{-7pt}	
	\resizebox{0.95\textwidth}{!}
	{
		\begin{tabular}{l|c||ccc||cc|cc|cc|cc||cc}
			\toprule
			&&\multicolumn{3}{c||}{Runtime} & \multicolumn{2}{c|}{07} & \multicolumn{2}{c|}{08} & \multicolumn{2}{c|}{09} &\multicolumn{2}{c||}{10} &\multicolumn{2}{c}{Mean on 07-10} \\ 
			\cline{3-15}\noalign{\smallskip}
			
			&\multirow{-2}{*}{\begin{tabular}[c]{@{}c@{}}Method \end{tabular}}
			& Data preparation & Inference & Total time& $t_{rel}$                          & $r_{rel}$   & $t_{rel}$ & $r_{rel}$ & $t_{rel}$                          & $r_{rel}$   & $t_{rel}$ & $r_{rel}$    & $t_{rel}$ & $r_{rel}$  

			\\
			\hline\hline
			\noalign{\smallskip}
			(a)&LO-Net + Mapping (with projection, all points) \cite{li2019net} &8.5 ms&71.6 ms &   80.1 ms  	&0.56 & 0.45 
			&\bf1.08 & 0.43 
			&\bf0.77 & 0.38 
			&0.92 &\bf0.41 
			&0.833  & 0.418   \\
			&Full LOAM (with mapping, all points) \cite{zhang2017low} &18.0 ms&51.0 ms &   69.0 ms  	&0.69 & 0.50 
			&1.18 & 0.44 
			&1.20 & 0.48 
			&1.51 & 0.57 
			& 1.145 &  0.498   \\
			&Ours (full, with projection, all points) & \bf 8.3 ms & \textbf{37.0 ms}   & \textbf{45.3 ms}     				&\bf0.46 & \bf0.38
			&1.14    &\bf0.41
			&0.78 & \bf0.33
			&\bf0.80 &0.46
			&\bf0.795  &\bf 0.395   \\
			\hline
			\noalign{\smallskip} 			
			(b)&Ours (w/o projection-aware grouping)	&\bf8.3 ms	&111.8 ms	&120.1 ms	&0.79	&0.62	&1.72	&0.80	&1.34	&0.53	&1.65	&0.70	&0.966	&0.505\\
&Ours (full, with projection-aware grouping)	&\bf8.3ms	&\bf37.0ms	&\bf45.3ms	&\bf0.46	&\bf0.38	&\bf1.14	&\bf0.41	&\bf0.78	&\bf0.33	&\bf0.80	&\bf0.46	&\bf0.795	&\bf0.395

			\\
			\hline
			\noalign{\smallskip}
			(c)&Ours (w/o filtering and with KNN selecting)& \bf 8.3 ms & 47.4 ms   & 55.7 ms      		&0.53 &\bf0.36
			&1.26 &0.48
			&0.85 &0.36
			&0.86 &\bf0.43
			&0.875 &0.408\\
			
			&Ours (full, with filtering and with random selecting)  & \bf8.3 ms & \textbf{37.0 ms}   & \textbf{45.3 ms}   			&\bf0.46 & 0.38
			&\bf1.14    &\bf0.41
			&\bf0.78 & \bf0.33
			&\bf0.80 & 0.46
			&\bf0.795  &\bf 0.395  \\ 
			\hline
			\noalign{\smallskip}
			(d) &Ours (w/o projection, with 8192 input points) \cite{wang2021pwclo}& \textbf{2.5 ms} & 63.9 ms   & 66.4 ms   	  		&0.60 & 0.44
			&1.26 & 0.55
			&0.79 & 0.35
			&1.69 &0.62
			&1.085  & 0.490  \\ 			
			&Ours (full, with projection, all points) &  8.3 ms & \textbf{37.0 ms}   & \textbf{45.3 ms}     				&\bf0.46 & \bf0.38
			&\bf1.14    &\bf0.41
			&\bf0.78 & \bf0.33
			&\bf0.80 &\bf0.46
			&\bf0.795  &\bf 0.395   \\

			\bottomrule
	\end{tabular}}
	\label{table:time}
\end{table*}

\subsection{Performance Evaluation}
As there are several modes to divide the training/testing sets for published papers \cite{li2019net,zheng2020lodonet,wang2019deeppco,zheng2020lodonet,cho2020}, in order to fairly compare with all current methods as we know, we train/test our model four times.

\begin{figure}[t]
	\centering
	\resizebox{1.00\columnwidth}{!}
	{
		\includegraphics[scale=1.00]{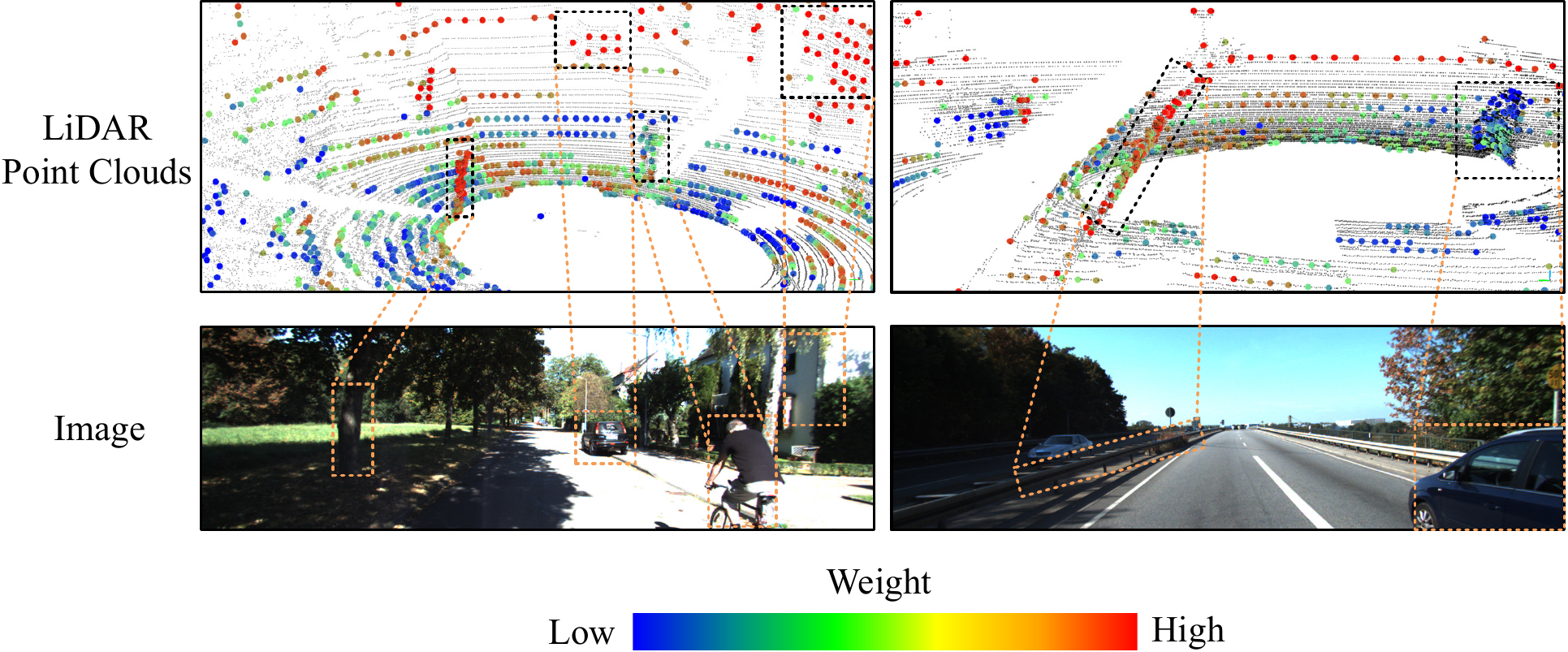}}
	\vspace{-14pt}
	\caption{The Visualization of embedding mask. In the LiDAR point clouds, the small red points are the whole point cloud of the first frame, and the big points with distinctive colors are the sampled points with contributions to the pose transformation. In the example on the left, the buildings and steel bars around the highway have high weights, while the bushes and the moving car have low weights. In the example on the right, the buildings, the tree trunks, and the static car have high weights, while the weeds and the cyclist have low weights.}
	\label{fig:visual}
\end{figure}

\vspace{5pt}
\noindent{}{\bf Using sequences 00-06/07-10 as training/testing sets:} Quantitative results are listed in Table~\ref{table:lidar}. ICP-point2point (ICP-po2po), ICP-point2plane (ICP-po2pl), GICP \cite{segal2009generalized}  , CLS \cite{velas2016collar} are several classic LiDAR odometry estimation methods based on Iterative Closest Point (ICP) \cite{besl1992method}. LOAM \cite{zhang2017low} is based on the matching of extracted line and plane features, which has a similar idea with our method, and it is a hand-crafted method. It achieves the best results among LiDAR-based methods in the KITTI odometry evaluation benchmark \cite{geiger2013vision}. Velas et al. \cite{velas2018cnn} is a learning-based method. It has a good performance when only the translation is estimated, but the performance will decrease when estimating the 6-DOF pose. LO-Net \cite{li2019net} and DMLO \cite{li2020dmlo} are learning-based LiDAR odometry methods that have comparable results, but they have no codes publicly available, so we adopt the same training and testing sequences with \cite{li2019net,li2020dmlo}. Compared with \cite{li2020dmlo}, our method utilizes soft correspondence rather than exacts matching pairs so as to realize end-to-end pose estimation. Compared with \cite{li2019net}, our method does not need an extra mask network and can filter the outliers with the proposed hierarchical embedding mask. Moreover, our method uses 3D feature learning ranther than 2D  \cite{li2019net,li2020dmlo} and obtains the LiDAR odometry from the raw 3D point clouds. We achieved better results than LO-Net~\cite{li2019net}, DMLO \cite{li2020dmlo}, even than LOAM~\cite{zhang2017low} on most sequences. We believe the pose refinement target makes our internal trainable mask effective for various outliers other than only dynamic regions~\cite{li2019net} as shown in Fig.~\ref{fig:visual}. Moreover, the proposed trainable iterative pose warp-refinement method refines the estimated pose multiple times in one end-to-end network inference,  contributing to the super performance.

\vspace{5pt}
\noindent{}{\bf Comparisons with HALFlow \cite{wang2021hierarchical}:} HALFlow \cite{wang2021hierarchical} is an end-to-end learnable 3D scene flow estimation method based on PWC structure. However, HALFlow \cite{wang2021hierarchical} only estimates the motion of each point. Thus, to generate a transformation of two point clouds, it needs further calculation, which is influenced a lot by the presence of dynamic obstacles and other outliers. So using HALFlow \cite{wang2021hierarchical} in odometry tasks will lead to performance degradation. Unlike HALFlow \cite{wang2021hierarchical}, we creatively design optimizable embedding masks and pose refinement modules to address the above challenges and successfully solve the odometry problem. The results in Table~\ref{table:wang} show the necessity and superior performance of our end-to-end trainable LiDRA odometry.

\vspace{5pt}
\noindent{}{\bf Using other sequences as training/testing sets: }

\noindent{}{\bf 00-06, 09-10/07-08:} Quantitative results are listed in Table~\ref{table:lidar_7_8} to compare with a recent method \cite{zheng2020lodonet}.

\noindent{}{\bf 00-03, 05-09/04, 10:} As shown in Table~\ref{table:lidar4_10}, we adopte the same training/testing sets of the KITTI odometry dataset \cite{geiger2013vision} and compare with \cite{wang2019deeppco}. 

\noindent{}{\bf 00-08/09-10:} Cho et al. \cite{cho2020} propose an unsupervised method on LiDAR odometry. Table~\ref{table:lidar09_10} shows the quantitative comparison of this method with ours.

The results in Table~\ref{table:lidar_7_8}, Table~\ref{table:lidar4_10}, and Table~\ref{table:lidar09_10} show that our method outperforms them. \cite{wang2019deeppco} is based on 2D convolutional network, which loses the raw 3D information. \cite{zheng2020lodonet} finds matched keypoint pairs in 2D depth images to regress pose. \cite{cho2020} is a method of unsupervised training. The results demonstrate the superiority of our full 3D learning-based LiDAR odometry. 

\vspace{5pt}
\noindent{}{\bf Using official test sequences 11-21 as testing sets: }
The official test sequences 11-21 of the KITTI odometry dataset is more difficult than the training and validation set with publicly available ground truth values. We evaluated our method and four classical geometry-based methods, pyLiDAR \cite{dellenbach2021s}, A-LOAM \cite{aloam}, LeGO-LOAM \cite{shan2018lego}, and SuMa \cite{behley2018efficient}, under the same conditions without mapping, on 11-21 sequences of the KITTI odometry dataset. For the results on sequences 11-21, our method loads the trained model on sequences 00-06 and then fine-tunes it using sequences 07-10. The experimental results are shown in Table \ref{table:11-21}. Although the more challenging environment does make all methods’ performance drop on the 11-21 sequence (including our method and all four geometry-based methods \cite{dellenbach2021s,aloam,shan2018lego,behley2018efficient}), our method still outperforms all four classical geometry-based methods at official test sequences 11-21. The experiments show the prospect and competitiveness of our method compared with geometric approaches.

\begin{figure}[t]
	\begin{center}
		\resizebox{1.00\columnwidth}{!}
		{
			\includegraphics[scale=1.00]{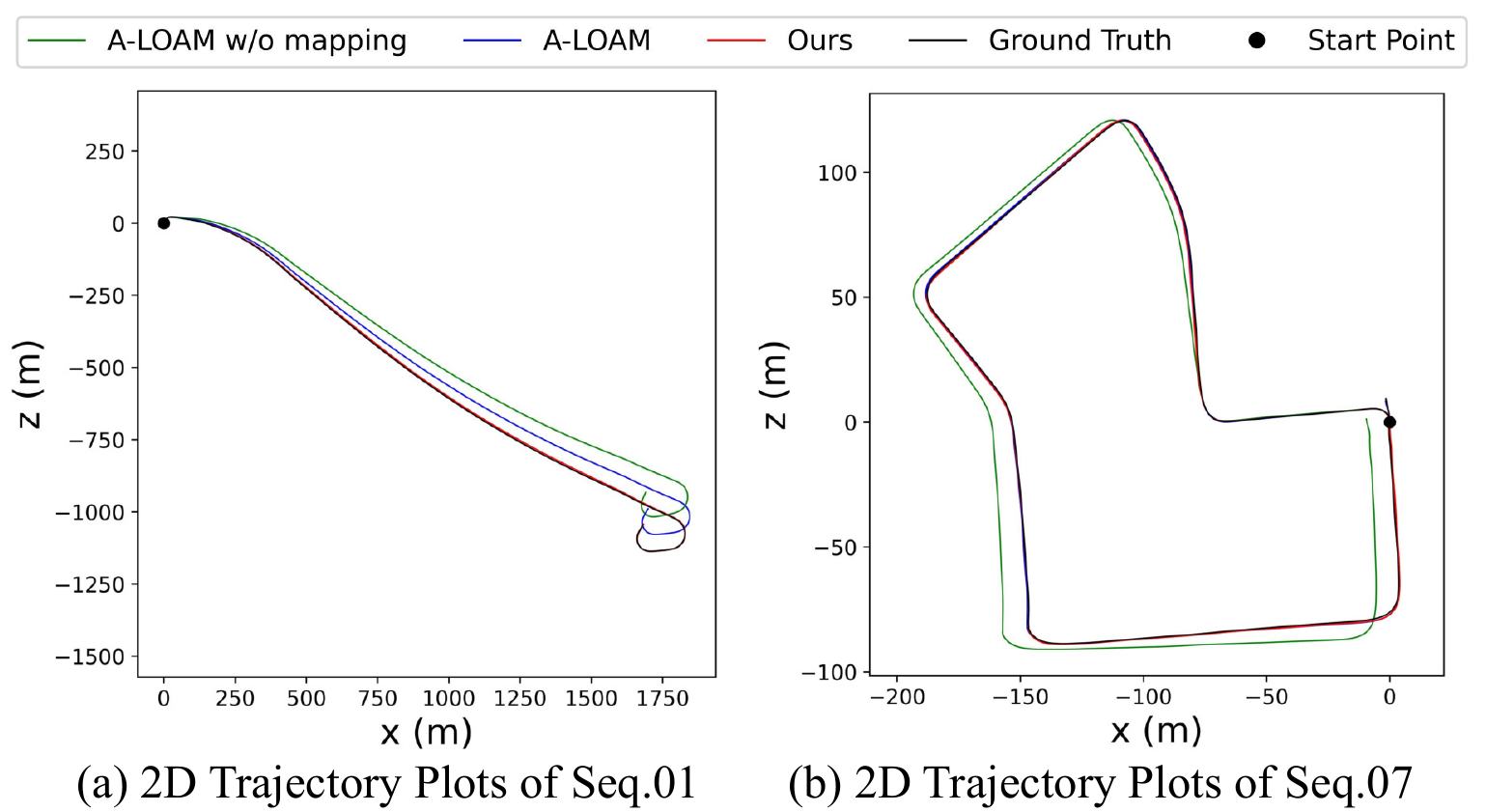}}
	\end{center}
	\vspace{-9pt}
	\caption{Trajectory results of A-LOAM and ours on KITTI training sequences with ground truth. Ours is much better than the A-LOAM without mapping. And ours also has similar performance on the two sequences with full A-LOAM, though ours is for odometry and A-LOAM has the mapping optimization.}
	\label{fig:odometry_path2d}
\end{figure}

\begin{figure}[t]
	\begin{center}
		\resizebox{1.00\columnwidth}{!}
		{
			\includegraphics[scale=1.00]{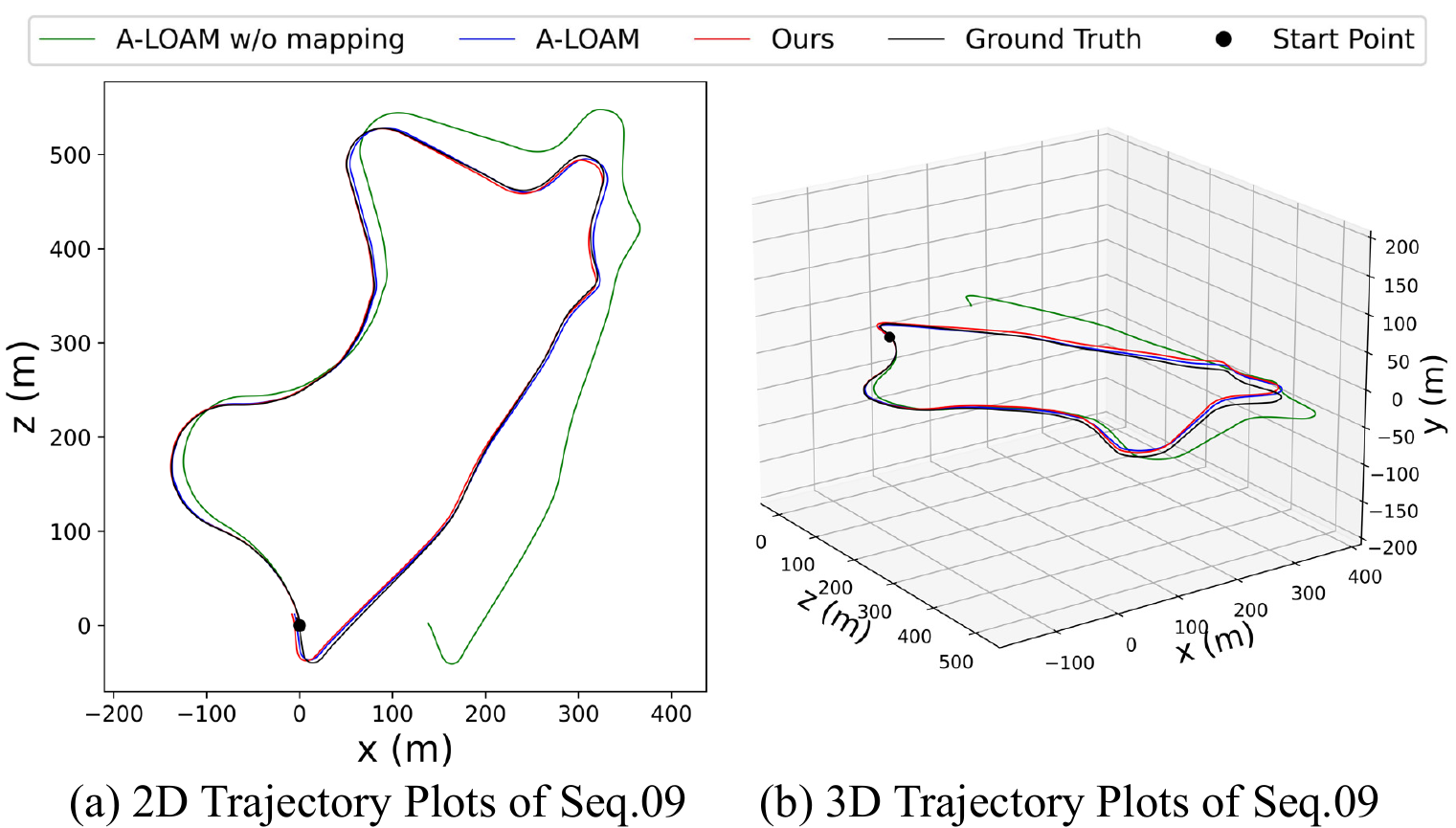}}
	\end{center}
	\vspace{-9pt}
	\caption{3D and 2D trajectory results on KITTI validation sequence 09 with ground truth. Our method obtained the similar accurate trajectory with full A-LOAM.}
	\label{fig:odometry_path2d3d}
\end{figure}

\begin{figure}[t]
	\begin{center}
		\resizebox{1.00\columnwidth}{!}
		{
			\includegraphics[scale=1.00]{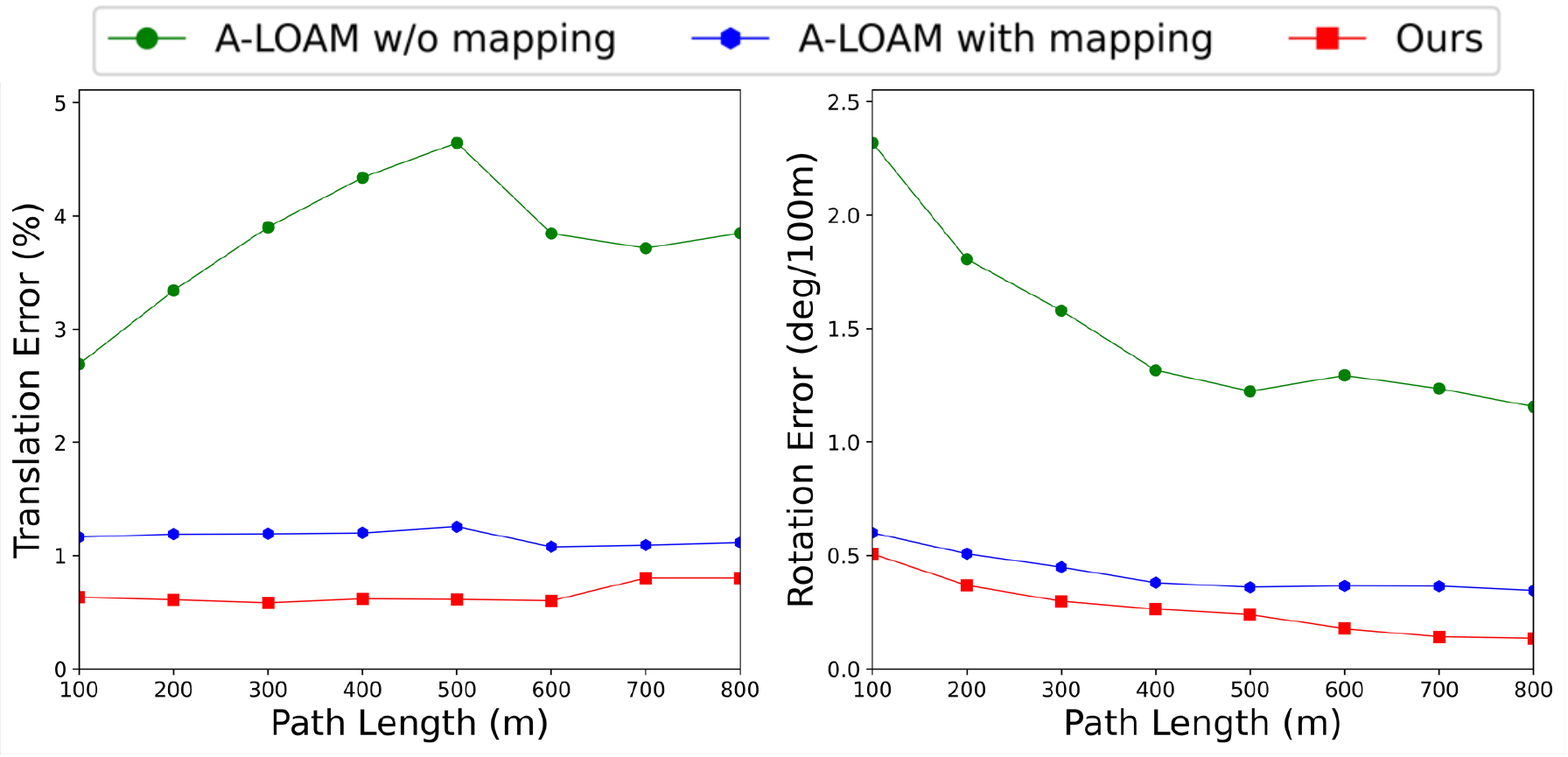}}
	\end{center}
	\vspace{-9pt}
	\caption{Average translational and rotational error on KITTI sequences 00-10 on all possible subsequences in the length of $100,200,...,800m$. Our method has the best performance.}
	\label{fig:odometry_error_line}
\end{figure}

{The qualitative results are shown in Figs.~\ref{fig:odometry_path2d},~\ref{fig:odometry_path2d3d}, and~\ref{fig:odometry_error_line}. We compared our method with A-LOAM \cite{aloam} and A-LOAM without mapping since \cite{li2019net,zheng2020lodonet,wang2019deeppco,li2020dmlo, zhang2017low} do not release their codes. A-LOAM \cite{aloam} is an advanced implementation of LOAM \cite{zhang2017low}. Full A-LOAM \cite{aloam} has a superb performance but degrades significantly without mapping. Ours is only for odometry and is better than A-LOAM without mapping. At the same time, ours is even better than A-LOAM \cite{aloam} on average evaluation.}

As shown in Table \ref{table:time}(a), we also compare the runtime with other methods that have published their efficiencies \cite{li2019net,zhang2017low}. Both LO-Net \cite{li2019net} and full LOAM \cite{zhang2017low} require mapping to achieve high accuracy, while our method is only for odometry and achieves both higher performance and efficiency compared to them. Note that the mapping and odometry of LOAM \cite{zhang2017low} are separate threads, so removing the mapping will not significantly reduce its time efficiency.
Finally, the proposed method in this paper can realize {20 Hz} real-time LiDAR odometry and has a higher performance than all previous learning-based methods and the geometry-based approach, LOAM with mapping optimization, to our knowledge.

\setlength{\tabcolsep}{0.9mm}
\begin{table*}[t]
	\footnotesize
	\caption{The ablation study results of LiDAR odometry for the network structure on KITTI odometry dataset \cite{geiger2013vision}.}
	\vspace{-18pt}	
	\begin{center}
		\resizebox{1.0\textwidth}{!}
		{
			\begin{tabular}{l|l||cc|cc|cc|cc|cc|cc|cc|cc|cc|cc|cc||cc}
				\toprule
				& &  \multicolumn{2}{c|}{00$^*$}  &\multicolumn{2}{c|}{01$^*$}      & \multicolumn{2}{c|}{02$^*$} & \multicolumn{2}{c|}{03$^*$} &  \multicolumn{2}{c|}{04$^*$} & \multicolumn{2}{c|}{05$^*$} & \multicolumn{2}{c|}{06$^*$} & \multicolumn{2}{c|}{07} & \multicolumn{2}{c|}{08} & \multicolumn{2}{c|}{09} &\multicolumn{2}{c||}{10} &\multicolumn{2}{c}{Mean on 07-10} \\ 
				\cline{3-26}\noalign{\smallskip}
				
				& \multirow{-2}{*}{\begin{tabular}[c]{@{}c@{}}Method \end{tabular}}
				&  $t_{rel}$  & $r_{rel}$   & $t_{rel}$                       & $r_{rel}$               & $t_{rel}$                          & $r_{rel}$   & $t_{rel}$ & $r_{rel}$   & $t_{rel}$                          & $r_{rel}$   & $t_{rel}$ & $r_{rel}$    & $t_{rel}$                          & $r_{rel}$   & $t_{rel}$ & $r_{rel}$ & $t_{rel}$                          & $r_{rel}$   & $t_{rel}$ & $r_{rel}$    & $t_{rel}$ & $r_{rel}$  & $t_{rel}$ & $r_{rel}$       \\
				\hline\hline
				\noalign{\smallskip}
				(a)   
				&Ours w/o mask    
				&0.70 & 0.31
				&0.80 & 0.28	
				& 0.72 & 0.28 
				& 0.79 &\bf0.38
				& 0.32 &  0.33
				& 0.80 &  0.39
				& 0.37 &  0.30
				& 0.91 & 0.51
				& 1.40 &  0.45
				& 1.23 &  0.43
				& 1.33 & 0.73 
				& 1.218 & 0.530  
				\\
				& Ours w/o mask optimazation   
				&\bf0.59& \bf0.29
				&0.59 & \bf0.21
				&0.76 & 0.26 
				&0.83 & 0.48
				&0.27 & 0.17
				&0.44  & 0.23
				&0.38 &\bf0.22
				&0.48 &\bf0.35
				&1.30 &  0.46
				&0.91 &  0.36
				&0.90 &\bf0.46
				&0.898  &0.408
				\\
				
				&Ours (full, with mask and mask optimazation)     
				&0.83 & 0.33
				&\bf0.55 & \bf0.21
				&\bf0.71 & \bf0.25
				&\bf0.49 & \bf0.38
				&\bf0.22 & \bf0.11
				&\bf0.34 & \bf0.21
				&\bf0.36 &0.24
				&\bf0.46 &0.38
				&\bf1.14    &\bf0.41
				&\bf0.78 & \bf0.33
				&\bf0.80 & \bf0.46
				&\bf0.795  &\bf 0.395      
				\\
				\cline{1-26}\noalign{\smallskip}

				(b) &Ours (w/o cost volume)    
				& 6.64 & 2.70
				& 4.90 & 1.68
				& 7.27  &  2.76
				& 4.49 &  3.36
				& 2.69 &  1.93
				&5.67 & 2.49
				& 4.14 & 2.11
				& 7.66 & 4.92
				& 19.11 &  6.97
				& 10.95 &  4.24
				& 9.44  &  4.46
				& 11.790  & 5.148
				\\
				
				&Ours (with the cost volume in \cite{liu2019flownet3d})    
				&\bf0.57 &\bf0.24
				&0.84 &0.25
				&\bf0.70  & 0.27
				
				&\bf0.47 &\bf0.35
				& 0.36 & 0.18
				& 0.51 & 0.26
				& 0.36 & 0.22
				& 0.63 & 0.40
				&\bf1.10 & 0.43
				&\bf0.75 &\bf0.31
				&1.00  & 0.53
				&0.870 & 0.418
				\\
				&Ours (with the cost volume in \cite{wu2019pointpwc})      
				&0.81 &0.38
				&0.72 & 0.26
				& 0.82 &  0.30
				& 0.66 &  0.55
				&0.37 & 0.22
				& 0.51 &  0.31
				&\bf0.33 & \bf0.21
				& 0.60 & \bf0.36
				& 1.17 & \bf0.36
				& 0.92 &  0.43
				&1.50  &0.74 
				& 1.048 &  0.473 
				\\
				&Ours (full, with the cost volume in \cite{wang2021hierarchical})     
				&0.83 &0.33
				&\bf0.55 & \bf0.21
				&0.71 & \bf0.25
				&0.49 &0.38
				&\bf0.22 & \bf0.11
				&\bf0.34 & \bf0.21
				&0.36 &0.24
				&\bf0.46 &0.38
				&1.14    &0.41
				&0.78 &0.33
				&\bf0.80 & \bf0.46
				&\bf0.795  &\bf 0.395      
				\\
				\cline{1-26}\noalign{\smallskip}

				(c) 
				&Ours w/o Pose Warp-Refinement    
				&3.25 & 1.43
				&2.45 & 0.73	
				&2.70 &1.28
				&2.60 &1.63
				&1.15 &2.89
				&2.86 &1.39
				&1.85 &0.84
				&1.87 &1.82
				&5.35 &2.44
				&4.63 &1.82
				&3.19 &1.88
				&3.76  &1.99 
				\\
				&Ours w/o Pose Warp    
				&2.23 & 1.04 
				&3.07 & 1.01 
				&2.13 & 0.83
				&3.07 & 1.53			
				&0.69 & 0.44		 
				&2.09 & 0.98		 
				&1.31 &0.62 
				&2.84 &2.37
				&4.28 &1.85
				&2.86 &1.40
				&2.60 &1.19
				&3.145 &1.703
				\\
				
				&Ours (full, with Pose Warp-Refinement)     
				&\bf0.83 & \bf0.33
				&\bf0.55 & \bf0.21
				&\bf0.71 & \bf0.25
				&\bf0.49 & \bf0.38
				&\bf0.22 & \bf0.11
				&\bf0.34 & \bf0.21
				&\bf0.36 & \bf0.24
				&\bf0.46 & \bf0.38
				&\bf1.14 &\bf0.41
				&\bf0.78 & \bf0.33
				&\bf0.80 & \bf0.46
				&\bf0.795  &\bf 0.395        
				\\
				
				\cline{1-26}\noalign{\smallskip}
				(d) 
				&Ours (first embedding on the last level)
				&\bf0.62 &\bf0.28 
				&\bf0.51 &\bf0.18
				&0.74 &0.30 
				&0.58 &0.37
				&\bf0.19 &0.23
				&0.51 &0.25
				&\bf0.29 &\bf0.16
				&0.68 &0.52
				&1.42 &0.59
				&1.41 &0.45
				&1.10  &0.62
				&1.153 &0.545
				\\
				
				&Ours (full, first embedding on the penultimate level)     
				&0.83 &0.33
				&0.55 &0.21
				&\bf0.71 & \bf0.25
				&\bf0.49 & \bf0.38
				&0.22 & \bf0.11
				&\bf0.34 & \bf0.21
				&0.36 &0.24
				&\bf0.46 & \bf0.38
				&\bf1.14    &\bf0.41
				&\bf0.78 & \bf0.33
				&\bf0.80 & \bf0.46
				&\bf0.795  &\bf 0.395     
				\\
				\cline{1-26}\noalign{\smallskip}
				(e) 
				&Ours (removing ground less than $0.55m$ in height)    
				&1.02 &0.57
				&1.25 &0.46
				&0.98 &0.50
				&0.97 &0.58
				&0.68 &0.53
				&0.80 &0.39
				&0.49 &0.48
				&0.84 &0.84
				&1.66 &0.67
				&1.54 &0.93
				&1.39 &0.97
				&1.358  &0.853   
				\\
				&Ours (removing ground less than $0.3m$ in height) 
				&0.98 &0.52
				&0.94 &0.32
				&1.01 &0.46
				&0.74 &0.64
				&0.41 &0.41
				&0.63 &0.36
				&0.58 &0.41
				&0.83 &0.79
				&2.06 &0.85
				&1.34 &0.61
				&1.57 &0.78
				&1.450  &0.758
				\\
				
				&Ours (full, reserving ground)     
				&\bf0.83 & \bf0.33
				&\bf0.55 & \bf0.21
				&\bf0.71 & \bf0.25
				&\bf0.49 & \bf0.38
				&\bf0.22 & \bf0.11
				&\bf0.34 & \bf0.21
				&\bf0.36 & \bf0.24
				&\bf0.46 & \bf0.38
				&\bf1.14    &\bf0.41
				&\bf0.78 & \bf0.33
				&\bf0.80 & \bf0.46
				&\bf0.795  &\bf 0.395    
				\\
				\bottomrule
			\end{tabular}
		}
	\end{center}
	\label{table:ablation1}
\end{table*}

\setlength{\tabcolsep}{0.9mm}
\begin{table*}[t]
	\footnotesize
	\caption{The ablation study results of LiDAR odometry for the projection-aware 3D feature learning on KITTI odometry dataset \cite{geiger2013vision}.}
	\vspace{-18pt}	
	\begin{center}
		\resizebox{1.0\textwidth}{!}
		{
			\begin{tabular}{l|l||cc|cc|cc|cc|cc|cc|cc|cc|cc|cc|cc||cc}
				\toprule
				& &  \multicolumn{2}{c|}{00$^*$}  &\multicolumn{2}{c|}{01$^*$}      & \multicolumn{2}{c|}{02$^*$} & \multicolumn{2}{c|}{03$^*$} &  \multicolumn{2}{c|}{04$^*$} & \multicolumn{2}{c|}{05$^*$} & \multicolumn{2}{c|}{06$^*$} & \multicolumn{2}{c|}{07} & \multicolumn{2}{c|}{08} & \multicolumn{2}{c|}{09} &\multicolumn{2}{c||}{10} &\multicolumn{2}{c}{Mean on 07-10} \\ 
				\cline{3-26}\noalign{\smallskip}
				
				& \multirow{-2}{*}{\begin{tabular}[c]{@{}c@{}}Method \end{tabular}}
				&  $t_{rel}$  & $r_{rel}$   & $t_{rel}$                       & $r_{rel}$               & $t_{rel}$                          & $r_{rel}$   & $t_{rel}$ & $r_{rel}$   & $t_{rel}$                          & $r_{rel}$   & $t_{rel}$ & $r_{rel}$    & $t_{rel}$                          & $r_{rel}$   & $t_{rel}$ & $r_{rel}$ & $t_{rel}$                          & $r_{rel}$   & $t_{rel}$ & $r_{rel}$    & $t_{rel}$ & $r_{rel}$  & $t_{rel}$ & $r_{rel}$       \\
				\hline\hline
				\noalign{\smallskip}

(a)
&Ours (with FPS)	&4.46	&2.16	&6.25	&2.88	&5.53	&2.30	&5.75	&2.71	&8.17	&4.50	&4.90	&2.39	&4.74	&2.29	&4.14	&2.97	&4.55	&2.02	&7.51	&2.69	&7.07	&3.43	&5.734	&2.758\\
&Ours (with random sampling)	&3.24	&1.66	&3.97	&1.59	&4.04	&1.74	&2.36	&1.95	&4.59	&3.42	&4.79	&1.91	&3.98	&1.66	&3.28	&2.21	&4.17	&1.59	&3.90	&1.68	&4.08	&1.76	&3.855	&1.925\\
&Ours (full, with stride-based sampling)	&\bf0.83	&\bf0.33	&\bf0.55	&\bf0.21	&\bf0.71	&\bf0.25	&\bf0.49	&\bf0.38	&\bf0.22	&\bf0.11	&\bf0.34	&\bf0.21	&\bf0.36	&\bf0.24	&\bf0.46	&\bf0.38	&\bf1.14	&\bf0.41	&\bf0.78	&\bf0.33	&\bf0.80	&\bf0.46	&\bf0.795	&\bf0.395

	\\
	\cline{1-26}\noalign{\smallskip}

(b)
&Ours (w/o projection-aware grouping)	&1.08	&0.48	&0.86	&0.29	&0.83	&0.33	&0.71	&0.86	&0.33	&0.24	&0.86	&0.46	&0.46	&\bf0.24	&0.79	&0.62	&1.72	&0.80	&1.34	&0.53	&1.65	&0.70	&0.966	&0.505\\
&Ours (full, with projection-aware grouping)	
&\bf0.83	&\bf0.33	&\bf0.55	&\bf0.21	&\bf0.71	&\bf0.25	&\bf0.49	&\bf0.38	&\bf0.22	&\bf0.11	&\bf0.34	&\bf0.21	&\bf0.36	&\bf0.24	

&\bf0.46	&\bf0.38	&\bf1.14	&\bf0.41	&\bf0.78	&\bf0.33	&\bf0.80	&\bf0.46	&\bf0.795	&\bf0.395
	\\
	\cline{1-26}\noalign{\smallskip}

				(c)
				&Ours (w/o 3D distance-based filtering)    
				&\bf0.70 &\bf0.30
				&0.74 &0.24
				&\bf0.66 &0.27
				&\bf0.49 &\bf0.32
				&0.24 &0.35
				&0.52 &0.25
				&\bf0.35 &\bf0.17
				&0.62 &0.51
				&1.40 &0.54
				&0.92 &\bf0.29
				&0.98 &0.56
				&0.980  &0.475  
				\\	
				&Ours (full, with 3D distance-based filtering)     
				&0.83 &0.33
				&\bf0.55 & \bf0.21
				&0.71 & \bf0.25
				&\bf0.49 &0.38
				&\bf0.22 & \bf0.11
				&\bf0.34 & \bf0.21
				&0.36 &0.24
				&\bf0.46 & \bf0.38
				&\bf1.14    &\bf0.41
				&\bf0.78 &0.33
				&\bf0.80 & \bf0.46
				&\bf0.795  &\bf 0.395    
				\\ 
				
				\cline{1-26}\noalign{\smallskip}
				(d) 
				&Ours (w/o filtering and with KNN selecting)    
				&0.58 &0.27
				&\bf0.51 &\bf0.18
				&\bf0.66 &\bf0.24
				&\bf0.39 &\bf0.29
				&\bf0.22 &0.14
				&0.54 &0.26
				&0.38 &0.26
				&0.53 &\bf0.36
				&1.26 &0.48
				&0.85 &0.36
				&0.86 &\bf0.43
				&0.875 &0.408  
				\\
				&Ours (with filtering and with KNN selecting) 
				&\bf0.56 &\bf0.25
				&0.59 &0.28
				&\bf0.66 &\bf0.24
				&0.57 &\bf0.29
				&0.23 &0.22
				&0.49 &0.25
				&\bf0.28 &\bf0.22
				&0.60 &0.41
				&1.40 &0.58
				&0.83 &\bf0.33
				&0.91 &0.49
				&0.935  &0.453
				\\
				&Ours (full, with filtering and with random selecting)           
				&0.83 &0.33
				&0.55 &0.21
				&0.71 &0.25
				&0.49 &0.38
				&\bf0.22 & \bf0.11
				&\bf0.34 & \bf0.21
				&0.36 &0.24
				&\bf0.46 &0.38
				&\bf1.14    &\bf0.41
				&\bf0.78 & \bf0.33
				&\bf0.80 &0.46
				&\bf0.795  &\bf 0.395    
				\\
				\cline{1-26}\noalign{\smallskip}
	(e)
	&Ours (w/o projection, with 8192 input points) \cite{wang2021pwclo}    
				&\bf0.78 & 0.42
				&0.67 &0.23
				&0.86 & 0.41
				&0.76 & 0.44
				&0.37 & 0.40
				&0.45 & 0.27
				&\bf0.27 &\bf0.22
				&0.60 & 0.44
				&1.26 & 0.55
				&0.79 & 0.35
				&1.69 &0.62
				&1.085  & 0.490 \\ 
&Ours (with 2D convolution, all points)	&3.74	&1.53	&2.82	&0.95	&3.01	&1.20	&3.82	&2.14	&2.21	&2.39	&4.20	&1.87	&2.33	&1.62	&6.22	&4.14	&6.11	&2.81	&4.05	&2.25	&4.43	&3.16	&3.904	&2.187\\
&Ours (full, with projection-aware 3D learning, all points)	&0.83	&\bf0.33	&\bf0.55	&\bf0.21	&\bf0.71	&\bf0.25	&\bf0.49	&\bf0.38	&\bf0.22	&\bf0.11	&\bf0.34	&\bf0.21	&0.36	&0.24	&\bf0.46	&\bf0.38	&\bf1.14	&\bf0.41	&\bf0.78	&\bf0.33	&\bf0.80	&\bf0.46	&\bf0.795	&\bf0.395\\

				\bottomrule
			\end{tabular}
		}
	\end{center}
	\label{table:ablation2}
\end{table*}

\subsection{Ablation Study}

In order to analyze the effectiveness of each module, we remove or change components of our model to do the ablation studies on KITTI odometry dataset. The training/testing details are the same as described in Sec.~\ref{sec:implementation}.

\vspace{5pt}
\noindent{}{\bf Benefits of Embedding Mask:}
We first remove the optimization of the mask, which means that the embedding masks are independently estimated at each level. Then we entirely remove the mask and apply the average pooling to embedding features to calculate the pose. The results in Table~\ref{table:ablation1}(a) show that the proposed embedding mask and its hierarchical optimization both contribute to better results. 

The mismatched objects are different in various scenes. In some scenes, the car is dynamic, while the car is static in another scene (Fig.~\ref{fig:visual}). Through the learnable embedding mask, the network learns to filter outlier points from the overall motion pattern. The network achieves the effect of RANSAC~\cite{fischler1981random} by only once network inference.

\vspace{5pt}
\noindent{}{\bf Different Cost Volume:} 
As there are different point feature embedding methods in point clouds, we compare three recent methods in our odometry task, including the flow embedding layer in FlowNet3D \cite{liu2019flownet3d}, point cost volume~\cite{wu2019pointpwc} and the attentive cost volume~\cite{wang2021hierarchical}. The results in Table~\ref{table:ablation1}(b) show that the model with double attentive cost volume has the best results out of  the three. Therefore, robust point association also contributes to the LiDAR odometry task.

\vspace{5pt}
\noindent{}{\bf Effect of Pose Warp-Refinement:}
We first remove the pose warping and reserve the hierarchical refinement of the embedding feature and the mask. Then we remove the full pose warp-refinement module, which means that the embedding features, mask, and pose are only calculated once. As a result, the performance degrades a lot for both as shown in Table~\ref{table:ablation1}(c), which demonstrates the importance of the coarse-to-fine refinement.

\vspace{5pt}
\noindent{}{\bf Suitable Level for the First Embedding:}
As fewer points have less structural information, it is needed to decide which level is used to firstly correlate the two point clouds and generate the embedding features. We test the different levels of the first feature embedding. The results in Table~\ref{table:ablation1}(d) demonstrate that the most suitable level for the first embedding is the penultimate level in the point feature pyramid, which shows that the 3D structure is also important for the coarse pose regression.

\begin{figure*}[t]
	\begin{center}
		\resizebox{1.0\textwidth}{!}
		{
			\begin{tabular}{cc}
				\subfigure{\includegraphics[width=0.5\linewidth]{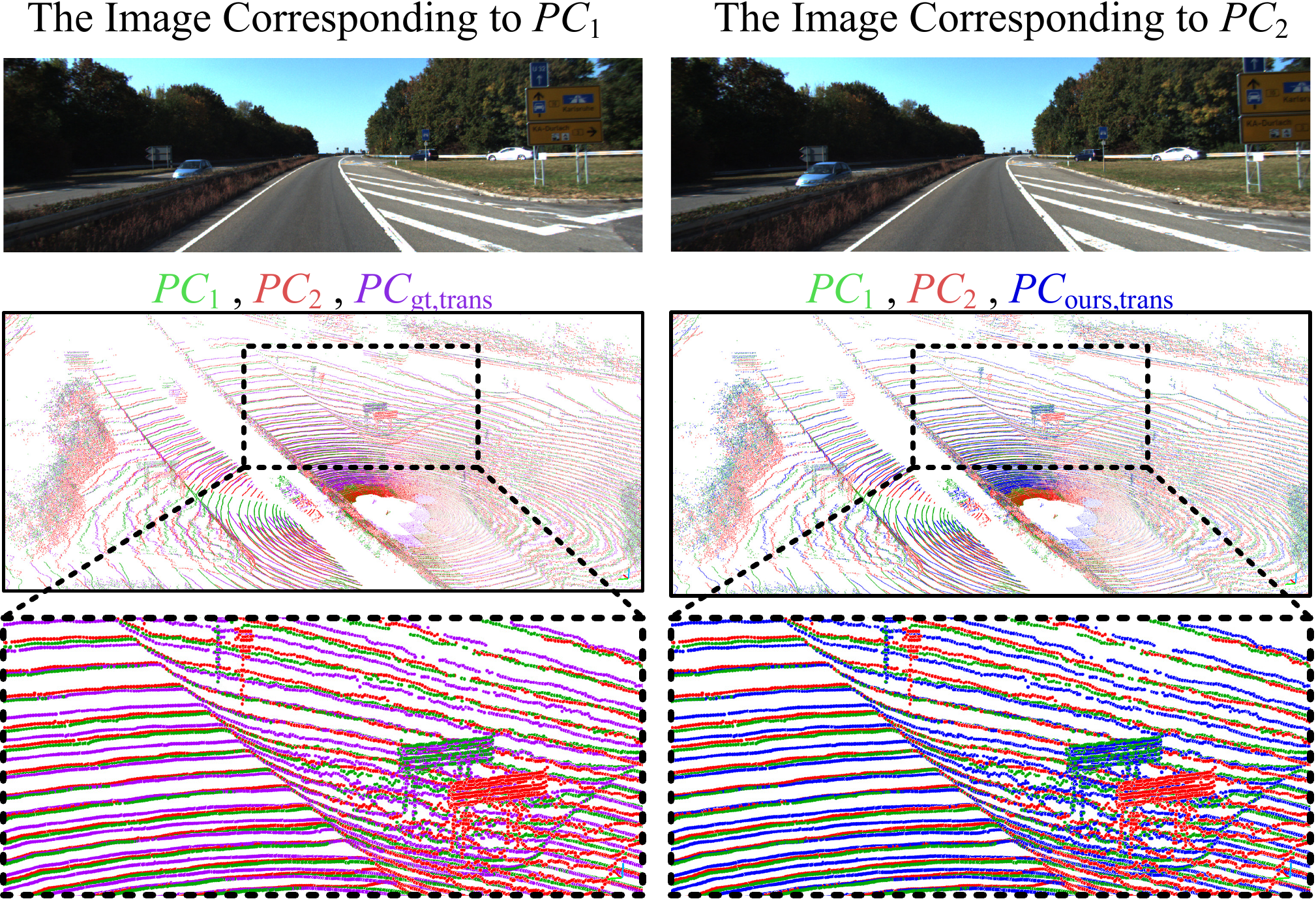}}& \subfigure{\includegraphics[width=0.5\linewidth]{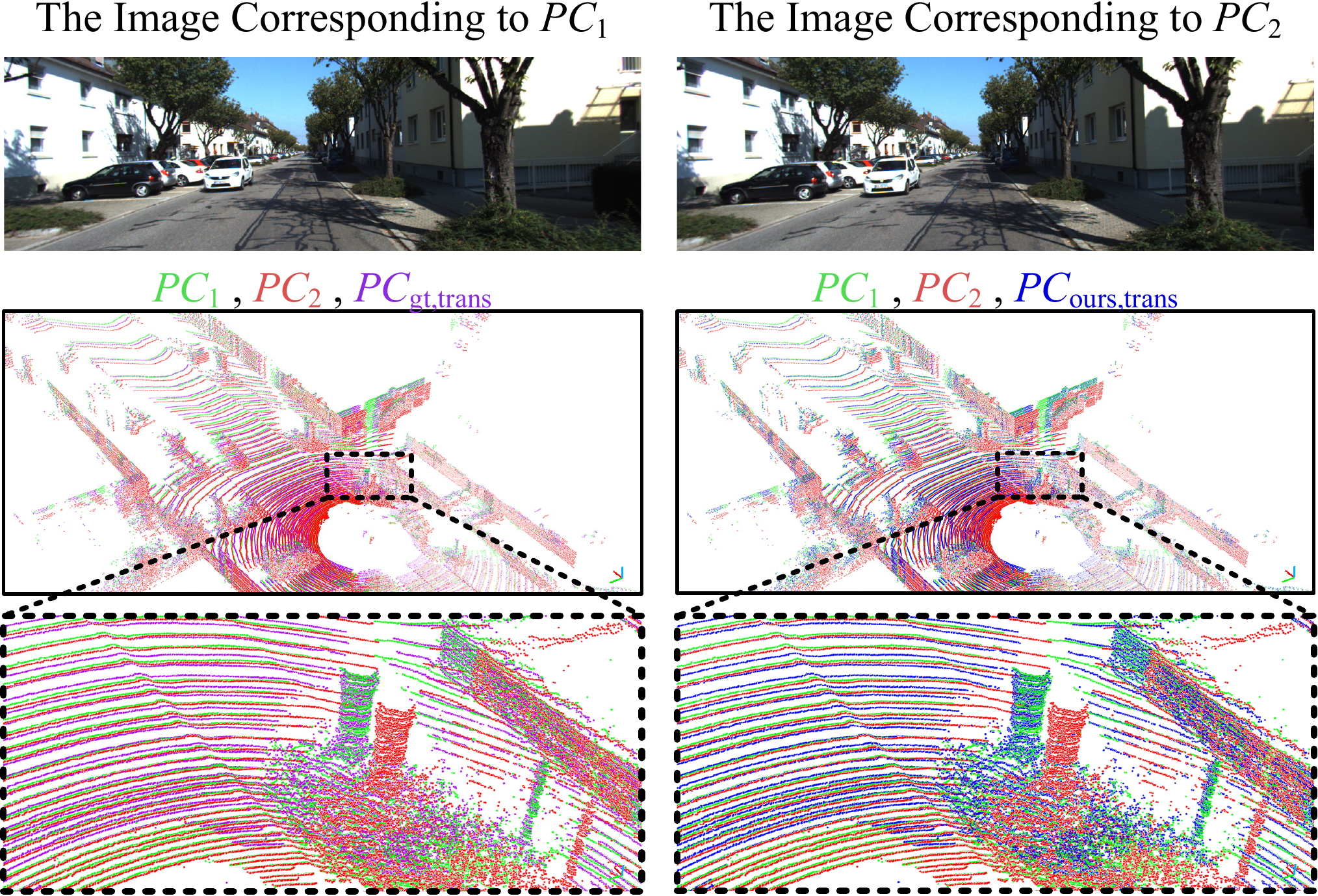}}
				\\
				{\small (a) Seq.01 Frames 167-168} &
				{\small (b) Seq.00 Frames 4463-4464} 
				\vspace{4pt}\\ 	
				\subfigure{\includegraphics[width=0.5\linewidth]{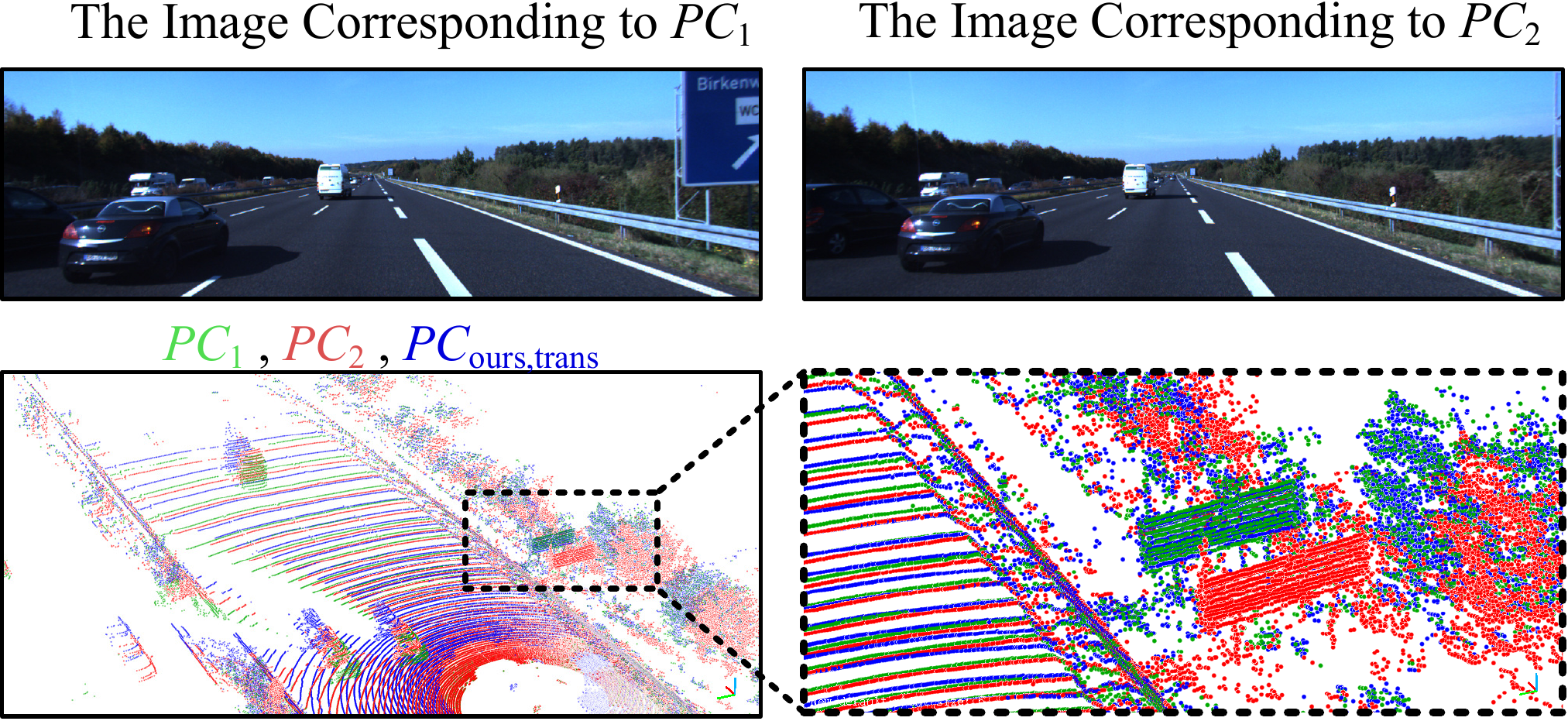}}&\subfigure{\includegraphics[width=0.5\linewidth]{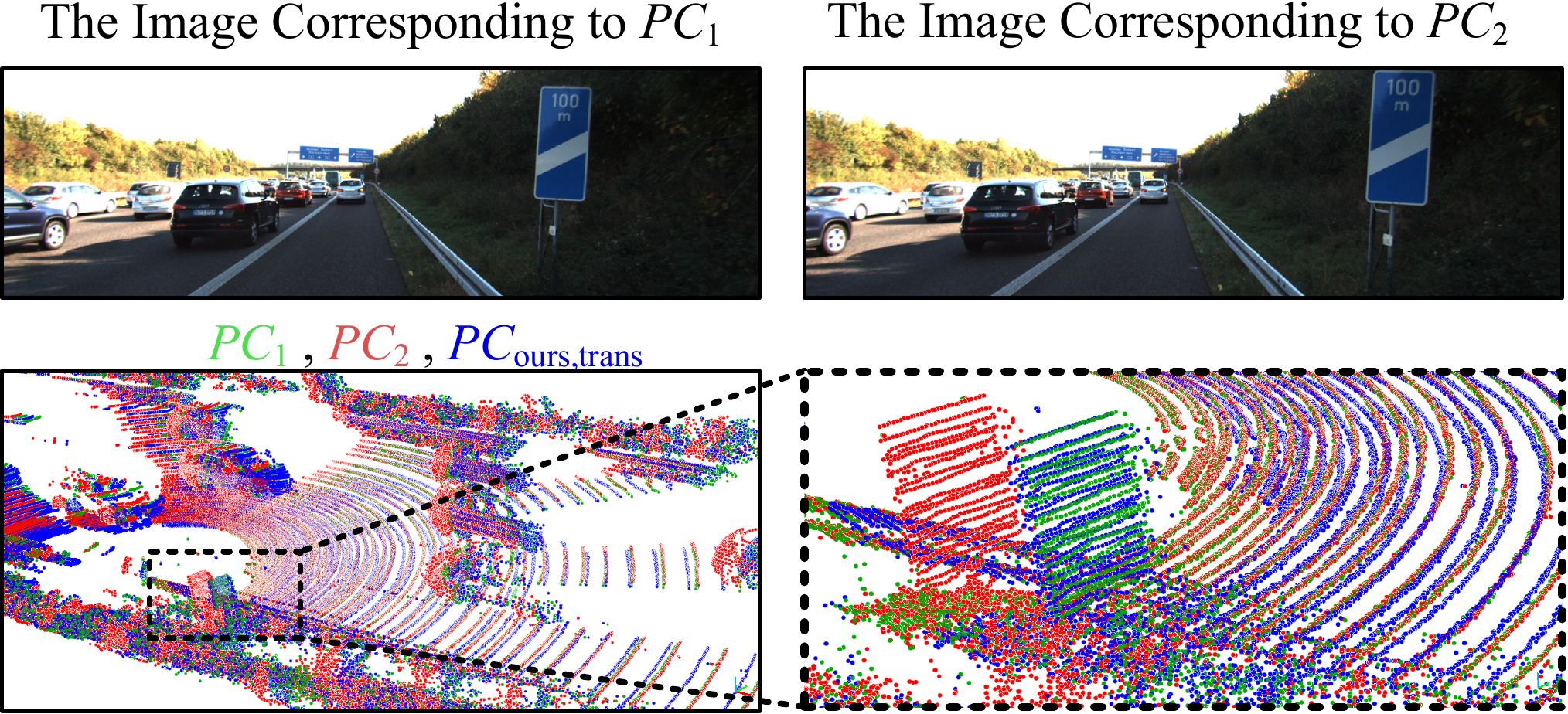}}
				
				\\{\small (c) Seq.21 Frames 841-842} &{\small (d) Seq.20 Frames 387-388} 
		\end{tabular}}
	\end{center}
	\vspace{-5pt}	
	\caption{\label{fig:dynimic_oboject} Cases to analyze the influence of dynamic objects on odometry estimation on KITTI odometry dataset~\cite{geiger2013vision}. The image corresponding to $PC_1$, the image corresponding to $PC_2$, the point clouds visulaization of $PC_1$ (green color), $PC_2$ (red color), $PC_{gt,trans}$ (purple color), and $PC_{ours,trans}$ (blue color) are visualized. We zoom in on some static details of the point clouds to see the registration effect as there are many dynamic objects in the scenes. The highly dynamic scenes in (c) and (d) have no ground truth.}
\end{figure*}

\vspace{5pt}
\noindent{}{\bf Removing or Reserving Ground:} We do the ablation study on removing the ground less than different heights, including $0.55m$ and $0.3m$. The comparison results in Table~\ref{table:ablation1}(e) show that the performance of reserving the ground is better due to the plane characteristics of the ground.

\vspace{5pt}
\noindent{}{\bf Stride-based Sampling:} We conduct ablation experiments by replacing the stride-based sampling used in the paper with other sampling methods. The FPS \cite{qi2017pointnet++} and random sampling \cite{hu2020randla} are the two common sampling methods. We replace the proposed stride-based sampling with these two sampling methods, while keeping the number of sampling points unchanged.  Both of these sampling methods degrade in results compared to our proposed stride-based sampling method as shown in Table \ref{table:ablation2}(a), which is because FPS and random sampling methods both lose the 2D ordered representation from the line scanning characteristics of the mechanical LiDAR after the sampling. The number of effective points is less when projecting the sampled points with FPS or random sampling to the 2D projection perception plane. Compared with these two sampling methods, our proposed stride-based sampling method in this paper maintains the 2D ordered representation after each sampling, so our proposed stride-based sampling gain the best results.

\vspace{5pt}
\noindent{}{\bf Projection-aware Grouping:} We conduct the ablation study of removing the projection-aware grouping operation, and the experimental results are in Table \ref{table:ablation2}(b). Experimental results show that similar performance is obtained without grouping. However, because there is no prior grouping operation, the subsequent filtering judgment will be performed on all points for each sampled central point, which will make the subsequent filtering operation computationally expensive. Therefore, we compared the efficiency of grouping and not grouping as shown in Table \ref{table:time}(b). Without grouping, the efficiency drops by a factor of nearly 10, which also reflects the importance of our proposed projection-aware grouping.

\vspace{5pt}
\noindent{}{\bf 3D Distane-based Filtering:} 
Since points that are close in cylindrical projection plane are not necessarily close in 3D, 3D distane-based filtering is used to filter out points that are far away. The ablation experiments in Table~\ref{table:ablation2}(c) demonstrate the effectiveness of  filtering distant points in 3D Euclidean space. By cylindrical projecting but retaining and processing the raw 3D coordinates, the high precision and efficiency of LiDAR odometry are realized at the same time. It is our advantage to retain the raw point cloud, because previous projection-based LiDAR odometry does not process the raw 3D point cloud coordinates and can not filter the distant points accurately.

\vspace{5pt}
\noindent{}{\bf KNN Selecting or Random Selecting:}
We also study the effect of KNN selecting and Random Selecting (RS) after the 3D distance-based filtering (which is similar to random selecting within a constrained ball) in the local point cloud grouping on the results. As shown in the experimental results in Table~\ref{table:ablation2}(d), for KNN, there is a better result without the distance-based filtering. Random sampling with the distance-based filtering has the result that exceeds KNN selecting with or without filtering. Especially on the training set, KNN has a good effect, but on the test set, the method of random selecting within a fixed distance has a better effect. Feature extraction within a ball with a fixed distance allows the network to learn 3D features with a fixed spatial scale, which brings stronger spatial generalization. In addition, because there is no need to sort when grouping, the random selecting method achieves speeding up by {10 ms} as shown in Table~\ref{table:time}(c).

\vspace{5pt}
\noindent{}{\bf Projection-aware 3D Feature Learning:}
Our previous conference version \cite{wang2021pwclo}, sampling 8192 raw 3D points per frame as network input has obtained SOTA performance that exceeds previous projection-based methods, which is enough to show the potential of the 3D point cloud based method in deep LiDAR odometry tasks. However, because of the efficiency problem of FPS and KNN, inputting more points will seriously reduce the efficiency. Therefore, this paper proposes projection prior based operators to accelerate global sampling and grouping operations of point cloud  in the network. The experimental results in Table~\ref{table:ablation2}(e) shows that the proposed projection-aware 3D method in this paper not only processes more points at once, but also achieves better performance. 
In addition, the runtime comparison in Table \ref{table:time}(d) shows that the proposed projection-aware 3D method achieves {20 ms} faster than the baseline method with unordered 3D point input. 

We also performed ablation experiments by replacing the aggregation with 2D convolution. The results in Table~\ref{table:ablation2}(e) show that our feature aggregation with 3D learning method gains better performance. PointNet++ \cite{qi2017pointnet++} has demonstrated the efficiency of MLP and max pooling to the raw 3D points. 2D convolution kernel is not good enough for raw 3D points because the 3D coordinates are discrete and uneven in each kernel.  Ours realized the high precision and efficiency of LiDAR odometry at the same time by the proposed projection-aware 3D feature learning method.

\subsection{Visualization of Embedding Mask}
We visualize the embedding mask in the last pose refinement layer to show the contribution of each point to the final pose transformation. As illustrated in Fig.~\ref{fig:visual}, the points that are sampled from the static and structured rigid objects such as buildings, steel bars, and the static car have higher weights. On the contrary, the points from the dynamic and irregular objects such as the moving car, the cyclist, bushes, and weeds have lower weights. In LO-Net~\cite{li2019net}, only the dynamic regions are masked, while our method also gives small weight to unreliable weeds and bushes. Therefore, the embedding mask can effectively reduce the influence of dynamic objects and other outliers on the estimation of pose transformation from adjacent frames. 
 
\subsection{Registration Visualization of Two Consecutive Frames}\label{sec:visulization}

We visualize the consecutive frames to present some successful and failing cases of pose estimation by registering the two point clouds through the ground truth pose and our estimated pose. We register the $PC_2$ to the first frame in the visualization. The registered $PC_{gt, trans}$ is obtained by:
\begin{equation}
	{PC_{gt, trans}} = {T_{gt}}{PC_2}\end{equation}
where ${T_{gt}}$ is the ground truth pose transformation between ${PC_1}$ and ${PC_2}$.
The registered $PC_{ours, trans}$ is obtained by:
\begin{equation}
	{PC_{ours, trans}} = {T_{ours}}{PC_2}\end{equation}
where ${T_{ours}}$ is the estimated pose transfrmation between ${PC_1}$ and ${PC_2}$.

The effect of registration is visualized in Fig. \ref{fig:dynimic_oboject}. 
In Fig.~\ref{fig:dynimic_oboject}, some highly dynamic scenarios are visualized to see the robustness of our model for dynamic objects. In Fig.~\ref{fig:dynimic_oboject}(c) and (d), we visualize two images in sequences 11-21 without ground truth because there are few highly dynamic scenarios in sequences 00-10 with ground truth. In Fig.~\ref{fig:dynimic_oboject}(a),(b), and (c), when there are many dynamic cars, ours also have the right correspondence between registered $PC_{ours, trans}$ and $PC_2$, which can be seen by zooming in on static objects. There is one case in Fig.~\ref{fig:dynimic_oboject}(d), where ours has a little error {because of a lot of dynamic objects} and few static, rigid objects.
Overall, the visualization of registration in many scenes demonstrates the effectiveness of our model for dynamics.

\subsection{Experiments on Other Datasets}\label{sec:otherdata}

We tested various other scenarios on the M2DGR dataset \cite{yin2021m2dgr}, which is a dataset for ground robots in different scenarios of Shanghai Jiao tong University (SJTU) campus, including the outdoor scenes, the hall, room, and the interior and exterior scenes of the lift obtained when the robot entering and exiting the lift. The LiDAR sensor of the robot is a Velodyne VLP-32C. 

We compare with four classical geometry-based methods, including pyLiDAR \cite{dellenbach2021s}, A-LOAM \cite{aloam}, LeGO-LOAM \cite{shan2018lego}, and SuMa \cite{behley2018efficient} without mapping in the various scenarios of this dataset, as these four geometry-based methods all have available reproducible codes. The experimental results are as shown in Table \ref{table:sjtu}. Experimental results show that our method performs well outdoors, in the hall, in the room, and in the scene of entering and exiting the lift. These scenarios are all different from autonomous driving scenarios, and our method achieves similar performance to pyLiDAR \cite{dellenbach2021s} without mapping, which demonstrates the generalization performance of our method in different scenarios.

 \setlength{\tabcolsep}{0.9mm}
 \begin{table}[t]
 	\centering
 	\footnotesize
 	\caption{The LiDAR odometry results on the testing sets of M2DGR dataset \cite{yin2021m2dgr}. As the dataset only has position ground truth and no angle ground truth, we only evaluate Absolute Trajectory Error (ATE) (m) \cite{sturm2012benchmark}.}
 	\vspace{-12pt}
 	\begin{center}
 		\resizebox{1.0\columnwidth}{!}
 		{
 			\begin{tabular}{l||c|c|c|c||c}
 				\toprule
 				&Outdoor01	&Hall05	&Room-dark06	&Lift04	&Mean \\ 
 				\cline{2-6}\noalign{\smallskip}
 				
 				\multirow{-2}{*}{\begin{tabular}[c]{@{}c@{}}Method \end{tabular}}
 				&ATE  & ATE  & ATE & ATE & ATE \\
 				\hline\hline
 				\noalign{\smallskip}
 				pyLiDAR (w/o mapping) &1.294	&\bf2.209	&0.333	&\bf1.504	&\bf1.335
 				\\ 
 				LeGO-LOAM (w/o mapping)	&3.711	&5.082	&0.871	&3.040	&3.176\\
 				
 				A-LOAM (w/o mapping)	&5.191	&3.152	&0.636	&1.749	&2.682\\
SuMa (w/o mapping)	&12.871	&4.070	&0.879	&Fail	&5.940
\\
 				Ours      
 				&\bf1.030	&2.534	&\bf0.314	&1.741	&1.405 
 				\\ \bottomrule
 			\end{tabular}
 		}
 	\end{center}
 	\vspace{-4pt}
 	\label{table:sjtu}
 \end{table}

In order to test the generalization of our method to different LiDAR configurations, we also perform the test on the Argoverse autonomous driving dataset \cite{chang2019argoverse}. Different from KITTI dataset, Argoverse dataset \cite{chang2019argoverse} contains LiDAR point cloud coming from two VLP-32 LiDAR sensors. The two LiDAR sensors have a 40$^{\circ}$ overlap in the vertical viewing angle. We use this dataset to validate the robustness of our approach to different LiDAR configurations across different datasets.

 \setlength{\tabcolsep}{1.8mm}
 \begin{table}[t]
 	\centering
 	\footnotesize
 	\caption{The LiDAR odometry results on sequences 00-23 of Argoverse dataset \cite{chang2019argoverse}. As the length of each sequence in the Argoverse dataset is short, and most of the lengths are within 100 m, so the KITTI odometer test metrics are no longer suitable for the Argoverse dataset. We employ Absolute Trajectory Error (ATE) (m) and Relative Pose Error (RPE) (m) \cite{sturm2012benchmark} to evaluate the performance of ours and other methods in Argoverse dataset \cite{chang2019argoverse}.}	
 	\vspace{-12pt}
 	\begin{center}
 		\resizebox{0.85\columnwidth}{!}
 		{
 			\begin{tabular}{l||c|c}
 				\toprule
 				&\multicolumn{2}{c}{Mean on test sequences 00-23} \\ 
 				\cline{2-3}\noalign{\smallskip}
 				
 				\multirow{-2}{*}{\begin{tabular}[c]{@{}c@{}}Method \end{tabular}}

 				& \makebox[0.10\textwidth][c]{ATE} &\makebox[0.10\textwidth][c]{RPE} \\
 				\hline\hline
 				\noalign{\smallskip}
 				pyLiDAR (w/o mapping)  
 				&6.900	&0.109 
 				\\ 
 				LeGO-LOAM (w/o mapping)  
 				&4.537	&0.110
 				\\ 
 				A-LOAM (w/o mapping) 
 				&4.138	&0.066 
 				\\ 
 				SuMa (w/o mapping)  
 				&3.663	&0.039
 				\\ 
 				Ours      
 				&\bf0.751&	\bf0.018
 				\\ \bottomrule
 			\end{tabular}
 		}
 	\end{center}
 	\vspace{-4pt}
 	\label{table:arg}
 \end{table}
We also compare with the four classical geometry-based methods, including pyLiDAR \cite{dellenbach2021s}, A-LOAM \cite{aloam}, LeGO-LOAM \cite{shan2018lego}, and SuMa \cite{behley2018efficient} without mapping. The experimental results are in Table \ref{table:arg}. Experiments show that our method outperforms the four classical geometry-based methods. The generalization performance of our method in different scenarios and different LiDAR configurations is demonstrated.

\section{Conclusion}
To the best of our knowledge, our method is the first efficient 3D deep LiDAR odometry, which makes our method a significantly different solution compared to previous work \cite{nicolai2016deep,velas2018cnn,wang2019deeppco,li2019net,li2020dmlo}. In our architecture, the projection-aware representation of the 3D point cloud and corresponding projection-aware 3D learning operations are proposed to achieve efficient learning of LiDAR odometry task from large-scale point clouds. PWC structure for this task is built for the first time to realize the hierarchical refinement of 3D LiDAR odometry in a coarse-to-fine approach. The hierarchical embedding mask optimization is proposed to deal with various outlier points. Through the end-to-end joint optimization of all all proposed modules, a new state-of-the-art (SOTA) deep LiDAR odometry solution is achieved with no need for an extra mask network \cite{li2019net}. 

Since our mask can filter a variety of outlier points that are not suitable for calculating overall motion, it is a direction worth exploring to combine our new mask with mapping optimization in the future. The projection-aware 3D data organization is also worth exploring for other real-time tasks, such as semantic segmentation and scene flow learning based on RGBD camera or LiDAR. 
 
\section*{Acknowledgments}
This work was supported in part by the Natural Science Foundation of China under Grant 62225309, Grant 62073222, Grant U21A20480, and Grant U1913204; in part by the Science and Technology Commission of Shanghai Municipality under Grant 21511101900; and in part by the Open Research Projects of Zhejiang Laboratory under Grant 2022NB0AB01.

\ifCLASSOPTIONcaptionsoff
  \newpage
\fi



\bibliographystyle{IEEEtran}
\bibliography{IEEEabrv,bare_jrnl}
%

%
\vspace{-10mm}
\begin{IEEEbiography}[{\includegraphics[width=1in,height=1.25in,clip,keepaspectratio]{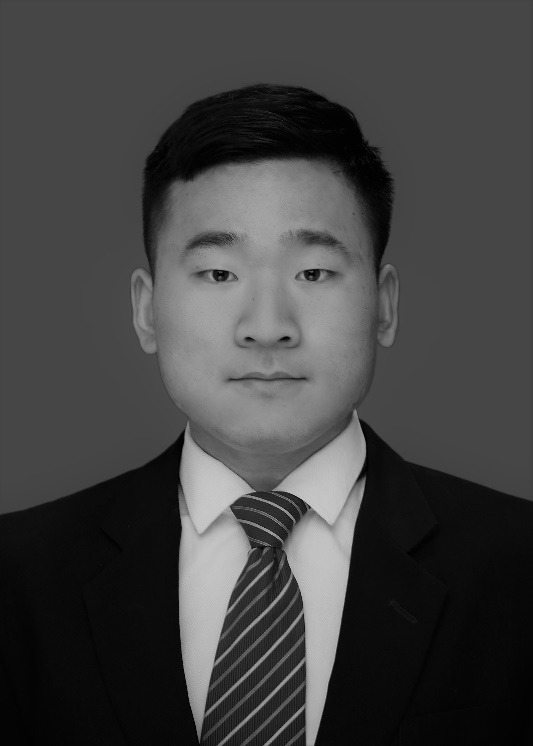}}]{Guangming Wang}
	received the B.S. degree from Department of Automation from Central South University, Changsha, China, in 2018. He is currently pursuing the Ph.D. degree in Control Science and Engineering with Shanghai Jiao Tong University. His current research interests include SLAM and computer vision, in particular, 3D scene flow estimation and LiDAR odometry.
\end{IEEEbiography}
\vspace{-10mm}
\begin{IEEEbiography}[{\includegraphics[width=0.9in,height=1.3in,clip]{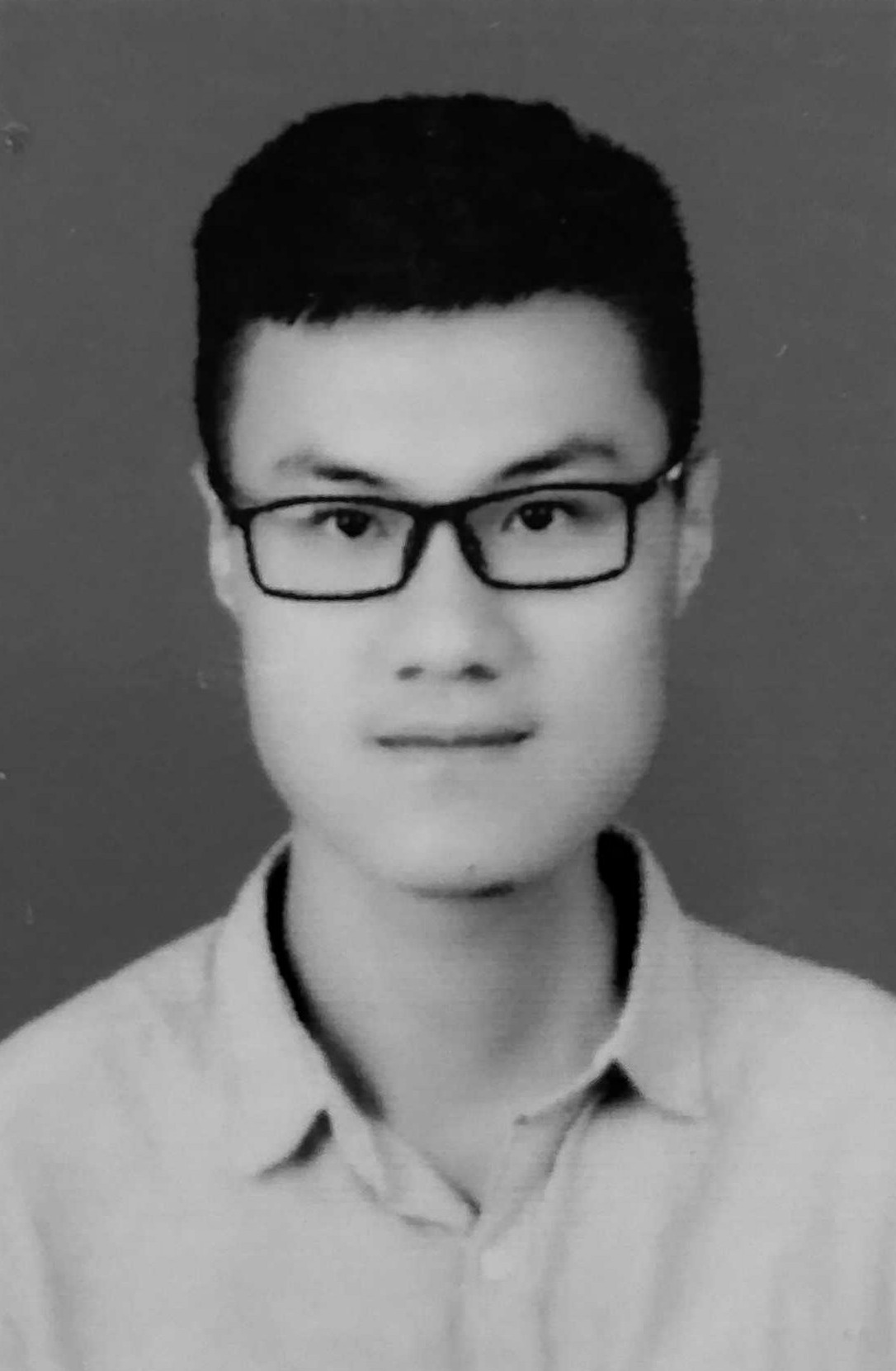}}]{Xinrui Wu} received the B.S. degree from the Department of Automation, Shanghai Jiao Tong University, Shanghai, China, in 2021, where he is currently pursuing the M.S. degree in Control Science and Engineering.
His latest research interests include SLAM and computer vision. 
\end{IEEEbiography}
\vspace{-10mm}
\begin{IEEEbiography}[{\includegraphics[width=1in,height=1.25in,clip]{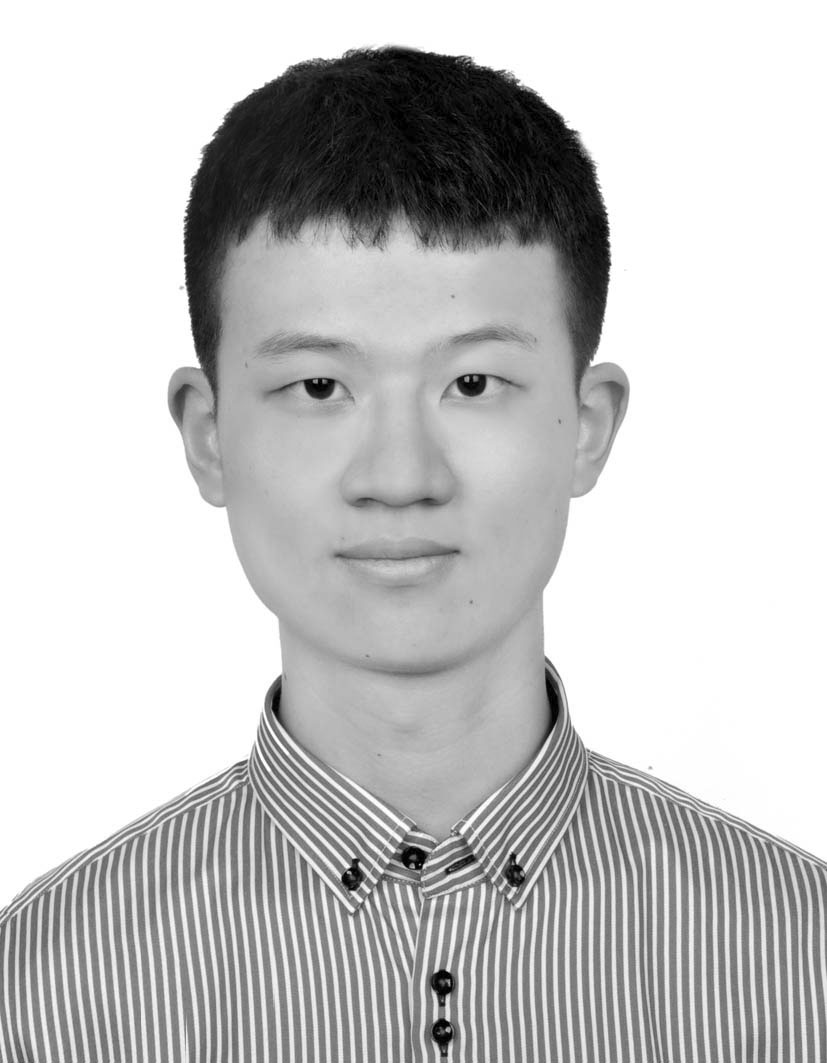}}]{Shuyang Jiang} is currently pursuing the B.S. degree in Department of Computer Science and Engineering (IEEE Honor Class), Shanghai Jiao Tong University. His latest research interests include SLAM and computer vision. 
\end{IEEEbiography}
\vspace{-10mm}
\begin{IEEEbiography}[{\includegraphics[width=1in,height=1.25in,clip,keepaspectratio]{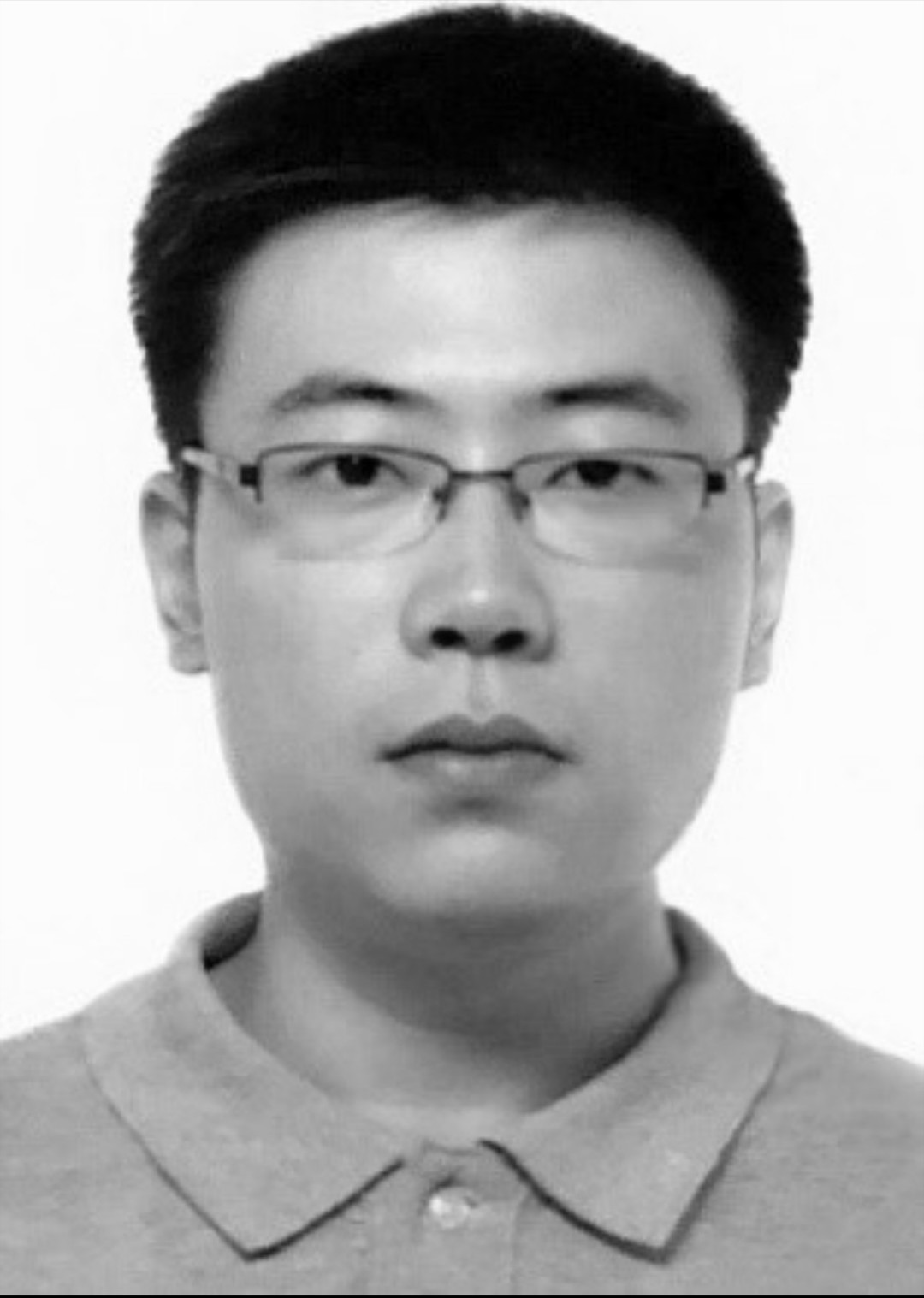}}]{Zhe Liu} received his B.S. degree in Automation from Tianjin University, Tianjin, China, in 2010, and Ph.D. degree in Control Technology and Control Engineering from Shanghai Jiao Tong University, Shanghai, China, in 2016. From 2017 to 2020, he was a Postdoctoral Fellow with the Department of Mechanical and Automation Engineering, The Chinese University of Hong Kong, Hong Kong. From 2020 to 2022, he was a Research Associate with the Department of Computer Science and Technology, University of Cambridge, Cambridge, U.K. He is currently an Associate Professor with the Key Laboratory of Artificial Intelligence of Ministry of Education, Shanghai Jiao Tong University.
His research interests include multirobot cooperation and autonomous
driving systems.
\end{IEEEbiography}
\vspace{-10mm}
\begin{IEEEbiography}[{\includegraphics[width=1in,height=1.25in,clip,keepaspectratio]{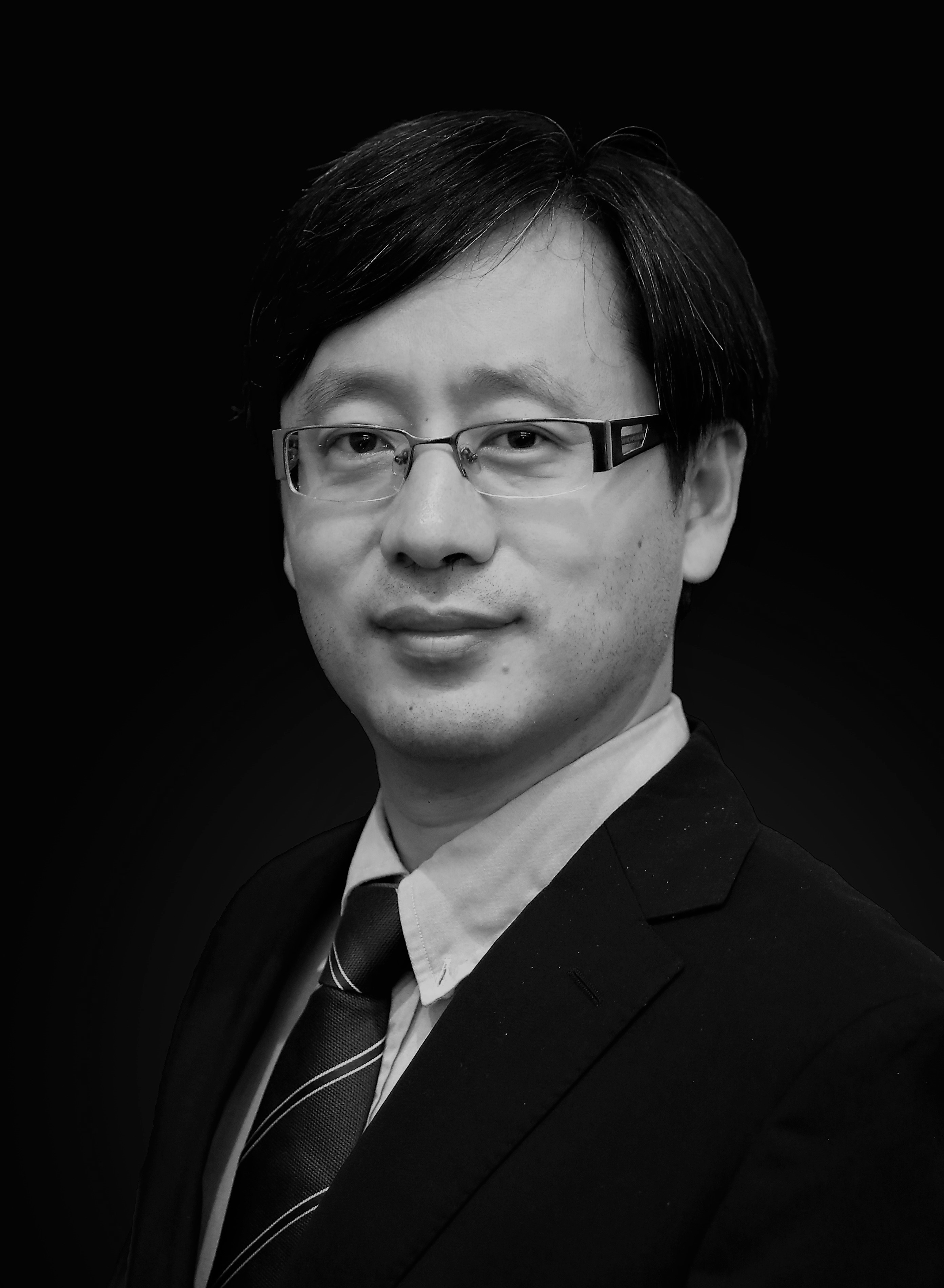}}]{Hesheng Wang} (Senior Member, IEEE) received the B.Eng. degree in electrical engineering from the Harbin Institute of Technology, Harbin, China, in 2002, and the M.Phil. and Ph.D. degrees in automation and computer-aided engineering from The Chinese University of Hong Kong, Hong Kong, in 2004 and 2007, respectively. He is currently a Professor with the Department of Automation, Shanghai Jiao Tong University, Shanghai, China. His current research interests include visual servoing, service robot, computer vision, and autonomous driving. Dr. Wang is an Associate Editor of IEEE Transactions on Automation Science and Engineering, IEEE Robotics and Automation Letters, Assembly Automation and the International Journal of Humanoid Robotics, a Technical Editor of the IEEE/ASME Transactions on Mechatronics, an Editor of Conference Editorial Board of IEEE Robotics and Automation Society. He served as an Associate Editor of the IEEE Transactions on Robotics from 2015 to 2019. He was the General Chair of IEEE ROBIO 2022 and IEEE RCAR 2016, and the Program Chair of the IEEE ROBIO 2014 and IEEE/ASME AIM 2019. He will be the General Chair of IEEE/RSJ IROS 2025. 
\end{IEEEbiography}

\end{document}